\documentclass[10pt,journal,compsoc]{IEEEtran}
\usepackage[table]{xcolor}
\usepackage{epsfig}
\usepackage{graphicx}
\usepackage{amsmath}
\usepackage{amssymb}
\usepackage{subfiles}

\usepackage{makecell} 
\usepackage{wrapfig} 
\usepackage{graphicx}
\usepackage[utf8]{inputenc} %
\usepackage[T1]{fontenc}    %
\usepackage{url}            %
\usepackage{amsfonts}       %
\usepackage{nicefrac}       %
\usepackage{microtype}      %
\usepackage{bm}
\usepackage[colorlinks=false]{hyperref}
\usepackage{cleveref}
\usepackage{array}
\usepackage{arydshln}
\usepackage{booktabs}
\usepackage{multirow}

\newcommand{\z}{{\rm\bf z}}         %
\newcommand{\x}{{\rm\bf x}}         %
\newcommand{\Loss}{\mathcal{L}}     %
\newcommand{\E}{\mathbb{E}}         %

\definecolor{ctext}{HTML}{228B8B}
\definecolor{caudio}{HTML}{3E4EB8}
\definecolor{cscene}{HTML}{E96E0E}

\newcommand{\ctext}[1]{\textcolor{ctext}{\hypersetup{citecolor=ctext}{#1}}}
\newcommand{\revision}[1]{\textcolor{black}{#1}}
\newcommand{\caudio}[1]{\textcolor{caudio}{\hypersetup{citecolor=caudio}{#1}}}
\newcommand{\cscene}[1]{\textcolor{cscene}{\hypersetup{citecolor=cscene}{#1}}}

\newlength\savewidth

\usepackage{xspace}

\makeatletter
\DeclareRobustCommand\onedot{\futurelet\@let@token\@onedot}
\def\@onedot{\ifx\@let@token.\else.\null\fi\xspace}

\def\eg{\emph{e.g}\onedot} 
\def\ie{\emph{i.e}\onedot} 
 
\def\etc{\emph{etc}\onedot} 
 
\def\etal{\emph{et al}\onedot}
\makeatother

\newcolumntype{S}{>{\centering\arraybackslash}m{0.9cm}}
\newcolumntype{M}{>{\centering\arraybackslash}m{1.2cm}}
\newcolumntype{L}{>{\centering\arraybackslash}m{1.4cm}}
\definecolor{mygray}{gray}{.95}
\definecolor{mylightergray}{gray}{.99}
\definecolor{mygreen}{RGB}{10, 179, 33}
\usepackage{caption} 
\renewcommand\arraystretch{1.15}
\makeatletter
\newcommand{\thickhline}{%
    \noalign {\ifnum 0=`}\fi \hrule height 1pt
    \futurelet \reserved@a \@xhline
}
\newcolumntype{"}{@{\vrule width 1pt}}

\ifCLASSOPTIONcompsoc
  \usepackage[nocompress]{cite}
\else
  \usepackage{cite}
\fi

\ifCLASSINFOpdf
\else
\fi

\begin{document}
\title{Human Motion Generation: A Survey}

\author{Wentao Zhu$^\ast$,
        Xiaoxuan Ma$^\ast$,
        Dongwoo Ro$^\ast$,
        Hai Ci,
        Jinlu Zhang, \\
        Jiaxin Shi,
        Feng Gao,
        Qi Tian,~\IEEEmembership{Fellow,~IEEE,}
        and Yizhou Wang
\IEEEcompsocitemizethanks{
\IEEEcompsocthanksitem Wentao Zhu, Xiaoxuan Ma, Dongwoo Ro, Hai Ci and Jinlu Zhang are with Center on Frontiers of Computing Studies, School of Computer Science, Peking University, Beijing 100871, China.
E-mail: \{wtzhu, maxiaoxuan, cihai\}@pku.edu.cn, \{dwro0121, jinluzhang\}@stu.pku.edu.cn.
\IEEEcompsocthanksitem Jiaxin Shi and Qi Tian are with Huawei Cloud Computing Technologies Co., Ltd., Shenzhen, Guangdong 518129, China.
E-mail: \{shijiaxin3, tian.qi1\}@huawei.com.
\IEEEcompsocthanksitem Feng Gao is with School of Arts, Peking University, Beijing 100871, China.
E-mail: gaof@pku.edu.cn.
\IEEEcompsocthanksitem Yizhou Wang is with Center on Frontiers of Computing Studies, School of Computer Science, Peking University, Beijing 100871, China, and also with Institute for Artificial Intelligence, Peking University, Beijing 100871, China.
E-mail: yizhou.wang@pku.edu.cn.
}

\thanks{
$\ast$ denotes equal contribution. \\
This work was supported in part by National Key R\&D Program of China (2022ZD0114900) and National Natural Science Foundation of China (General Program, No. 62176006). \\
(Corresponding authors: Feng Gao, Qi Tian, and Yizhou Wang.)
}}

\IEEEtitleabstractindextext{%
\begin{abstract}

Human motion generation aims to generate natural human pose sequences and shows immense potential for real-world applications. Substantial progress has been made recently in motion data collection technologies and generation methods, laying the foundation for increasing interest in human motion generation. Most research within this field focuses on generating human motions based on conditional signals, such as text, audio, and scene contexts. While significant advancements have been made in recent years, the task continues to pose challenges due to the intricate nature of human motion and its implicit relationship with conditional signals. In this survey, we present a comprehensive literature review of human motion generation, which, to the best of our knowledge, is the first of its kind in this field. We begin by introducing the background of human motion and generative models, followed by an examination of representative methods for three mainstream sub-tasks: text-conditioned, audio-conditioned, and scene-conditioned human motion generation. Additionally, we provide an overview of common datasets and evaluation metrics. Lastly, we discuss open problems and outline potential future research directions. We hope that this survey could provide the community with a comprehensive glimpse of this rapidly evolving field and inspire novel ideas that address the outstanding challenges.

\end{abstract}

}

\maketitle

\IEEEdisplaynontitleabstractindextext

\IEEEpeerreviewmaketitle

\IEEEraisesectionheading{\section{Introduction}\label{sec:introduction}}
\IEEEPARstart{H}umans plan and execute body motions based on their intention and the environmental stimulus~\cite{BernhardHommel1997TowardAA, blakemore2001perception}. As an essential goal of artificial intelligence, generating human-like motion patterns has gained increasing interest from various research communities, including computer vision~\cite{Guo_2022_CVPR,Kim_2022_CVPR}, computer graphics~\cite{alexanderson2022listen, Ao2023GestureDiffuCLIP}, multimedia~\cite{guo2020action2motion,PC-Dance}, robotics~\cite{nishimura2020long, gulletta2020human}, and human-computer interaction~\cite{Aud2Repr2Pose, yin2022one}. The goal of human motion generation is to generate natural, realistic and diverse human motions that can be used for a wide range of applications, including film production, video games, AR/VR, human-robot interaction, and digital humans. 

\begin{figure*}[t]
  \centering
  \includegraphics[width=0.8\linewidth]{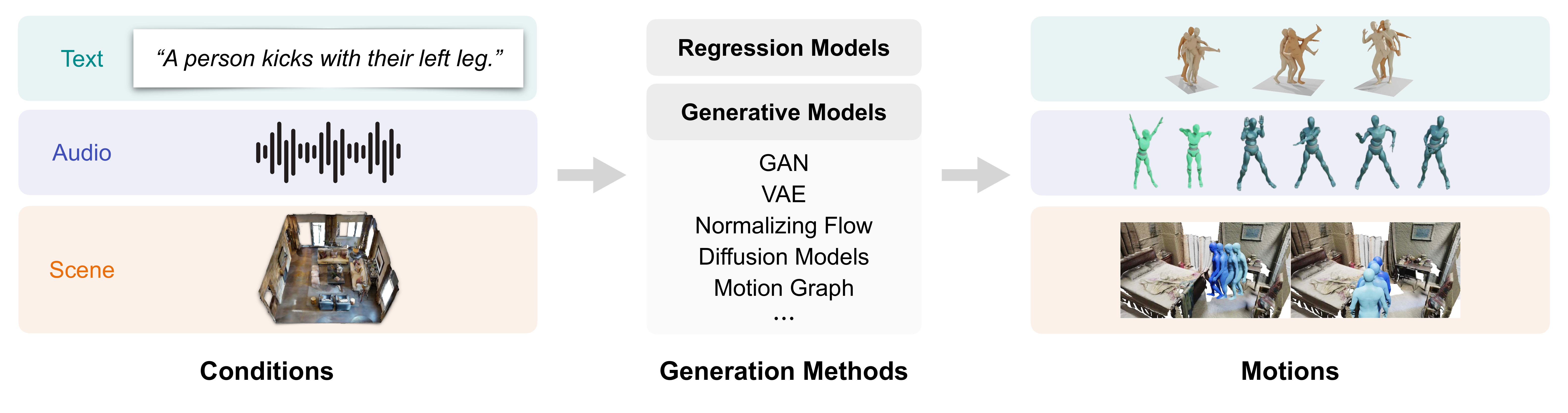}
  \vspace{-0.4cm}
  \caption{An overview of typical human motion generation approaches. Example images adapted from ~\cite{chang2017matterport3d,tevet2022human,tseng2022edge,wang2022humanise}.}
    \label{fig:pipeline}
\end{figure*}

\begin{figure*}[t]
  \vspace{-0.4cm}
  \centering
  \includegraphics[width=0.95\linewidth]{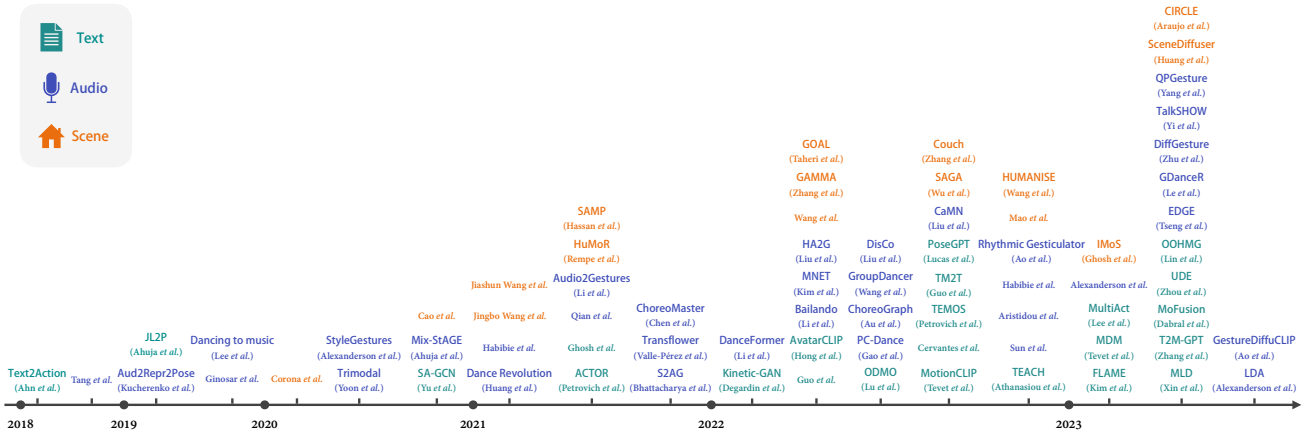}
  \caption{Recent advances of human motion generation methods with different conditions. }
    \vspace{-0.4cm}
  \label{fig:timeline}
\end{figure*}

With the rise of deep learning~\cite{lecun2015deep}, recent years have witnessed a rapid development of various generation methods, \eg, Autoregressive models~\cite{AR}, Variational Autoencoders (VAE)~\cite{vae}, Normalizing Flows~\cite{normalizingflows}, Generative Adversarial Networks (GAN)~\cite{gan}, and Denoising Diffusion Probabilistic Models (DDPM)~\cite{DDPM}. These methods have demonstrated great success across different domains, including text~\cite{gpt2020, ouyang2022training}, image~\cite{stylegan,stylegan2,stylegan3}, video~\cite{ho2022video, stylegan-v, digan}, and 3D objects~\cite{poole2023dreamfusion, gao2022get3d}. On the other hand, the remarkable progress in human modeling~\cite{loper2015smpl, SMPL-X:2019, MANO:SIGGRAPHASIA:2017} makes it easier to extract human motion from videos~\cite{martinez_2017_3dbaseline,pavllo20193d,kocabas2020vibe} and construct large-scale human motion datasets~\cite{h36m_pami,AMASS:2019,HUMBI,cai2022humman}. Consequently, the community has gained increasing interest in data-driven human motion generation over the past few years.

Nonetheless, human motion generation presents a complex challenge that extends beyond the mere application of deep generative models to human motion datasets. First, human motion is highly non-linear and articulated, subject to physical and bio-mechanical constraints. Additionally, human brains possess specialized neural mechanisms for perceiving biological motion~\cite{blakemore2001perception, grossman2000brain} and are sensitive to even slightly unnatural kinematics~\cite{troje2002decomposing, shimada2012modulation}. As a result, high visual quality is required for generated motions in terms of naturalness, smoothness, and plausibility. 
Second, the demand for human motion generation often includes a context as the conditional signal, such as text description, background audio, or surrounding environments, as shown in Figure ~\ref{fig:pipeline}. Generated motion should not only be plausible in itself but also harmonious with the conditional signal.
Third, human motion serves as an essential nonverbal communication medium, reflecting various underlying factors, such as goals, personal styles, social norms, and cultural expressions~\cite{MotionandRepresentation}. Ideally, motion generation models should learn to capture subtle variations and the semantic connection with the conditional signals.

In light of the rapid development and emerging challenges, we present a comprehensive survey of this field to help the community keep track of its progress. 
In Figure ~\ref{fig:timeline}, we summarize the development of human motion generation methods in recent years.
The rest of the survey is organized as follows. In Section~\ref{sec:scope}, we discuss the scope of this survey. Section~\ref{sec:preliminaries} covers the fundamentals of the task, including representations of human motion, motion data collection techniques, and various generative methods. In Section~\ref{sec:text2motion},~\ref{sec:audio2motion}, and ~\ref{sec:scene2motion}, we summarize existing approaches for human motion generation based on different conditional signals respectively, including text, audio, and scene. Section~\ref{sec:dataset} introduces the commonly used datasets and their properties. Section~\ref{sec:metrics} summarizes the evaluation metrics from various perspectives. Finally, we draw conclusions and provide some future directions for this field in Section ~\ref{sec:conclusion}.

\vspace{-0.17cm}

\section{Scope}
\label{sec:scope}

This survey focuses on the generation of human motion based on given conditional signals. We primarily discuss text, audio, and scene conditions. \revision{Some works also propose to generate human motion based on other conditions (\eg, others' motion ~\cite{joo2019towards}).}
With respect to the generation target, we incorporate different types of human motion representations, such as sequences of 2D/3D body keypoint, joint rotations, and parametric human body models~\cite{loper2015smpl, SMPL-X:2019}. 
We do not cover methods on human motion completion (\eg, motion prediction, motion interpolation) or human motion editing (\eg motion retargeting, motion style transfer). For reviews on these approaches, we direct readers to~\cite{lyu20223d, rudenko2020human, haarbach2018survey, akber2023deep}.
Additionally, we do not discuss works on generating human motion using physical simulation environments (\eg, character control, locomotion); please refer to~\cite{mourot2022survey} for a summary of such methods.
This survey serves as a complement to existing survey papers that focus on human pose estimation~\cite{chen2020monocular, liu2022recent}, motion capture~\cite{moeslund2001survey, moeslund2006survey}, and deep generative models~\cite{bond2021deep, suzuki2022survey, shi2022deep}.

\section{Preliminaries}
\label{sec:preliminaries}

\subsection{Motion Data}
We first introduce the human motion data representations, then discuss various human motion data collection techniques and their characteristics. 

\subsubsection{Motion Data Representation}

Human motion data can be effectively represented by the sequence of human body poses over the temporal dimension. More specifically, we categorize the data representations into \emph{keypoint-based} and \emph{rotation-based}. \revision{It is worth noting that a conversion is possible between these two types of representations. We can transition from joint rotations to keypoints using forward kinematics (FK), and inversely, from keypoints to joint rotations using {inverse kinematics (IK)}.}

\begin{figure}
    \centering
    \includegraphics[width=\linewidth]{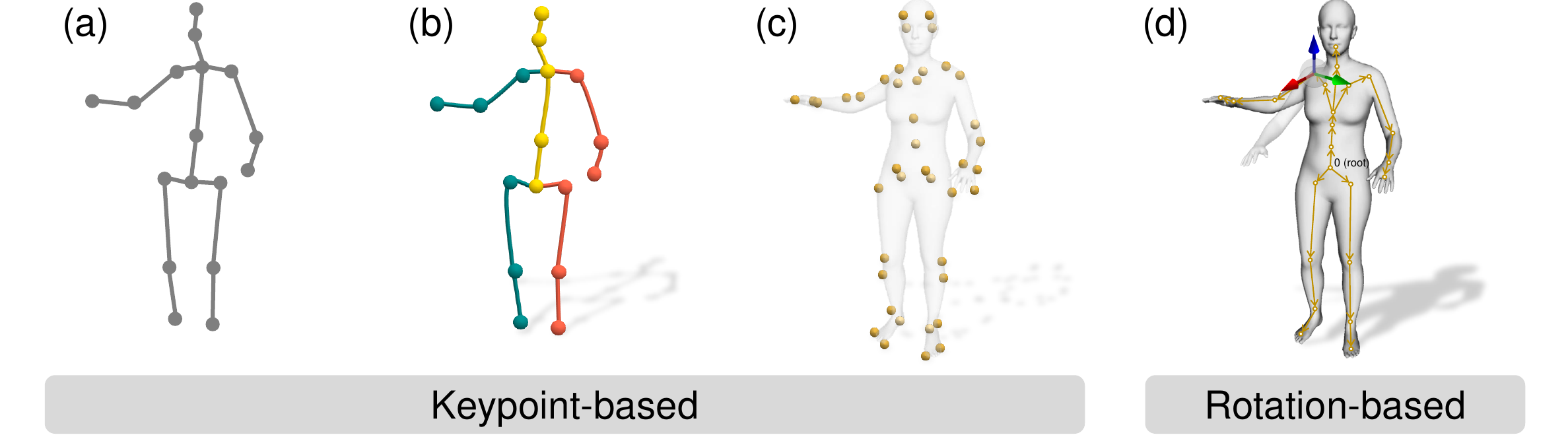}
    \caption{Typical human pose and shape representations with the same pose in (a) 2D keypoints, (b) 3D keypoints, (c) 3D marker keypoints, and (d) rotation-based model. }
    \vspace{-3mm}
    \label{fig:pose_rep}
\end{figure}

\noindent\textbf{Keypoint-based.} The human body is represented by a set of keypoints, which are specific points on the body that correspond to anatomical landmarks, such as joints or other significant locations. Each keypoint is represented by its 2D/3D coordinates in the pixel or world coordinate system, as shown in Figure \ref{fig:pose_rep} (a) (b). The motion data is then represented as a sequence of keypoint configurations over time. Keypoint-based representations can be directly derived from motion capture systems and exhibit great interpretability. However, in order to use keypoint-based data for animation or robotics, it is usually necessary to solve the inverse kinematics (IK) problem and convert the keypoints to rotations. Recently, some work \cite{zhang2021we, zanfir2021thundr, ma20233d} propose to represent the human pose with more landmarks on the surface of the human body, \ie, body markers, as shown in Figure \ref{fig:pose_rep} (c). Compared to traditional skeleton keypoints, body markers provide more comprehensive information in terms of body shapes and limb twists.

\noindent\textbf{Rotation-based.} Human pose could also be represented by joint angles, \ie, the rotation of the body parts or segments, relative to their parent in a hierarchical structure. Most studies consider 3D joint rotations in $\mathrm{SO}(3)$ and the rotations can be parameterized using various formats, such as Euler angles, axis angles, and quaternions. Based on the joint angle, some works \cite{loper2015smpl, SMPL-X:2019} model the human with statistical mesh models that further capture the shape of the body and the deformations that occur during movement. A widely-used statistical body model is the Skinned Multi-Person Linear (SMPL) model \cite{loper2015smpl}.

The SMPL model is parametrized by a set of pose and shape parameters, which can be used to generate a 3D mesh of a human body in a specific pose and shape, as shown in Figure~\ref{fig:pose_rep} (d). Pose parameters $\boldsymbol{\theta} \in \mathbb{R}^{K \times 3}$ of each joint are defined by the relative rotation with respect to its parent in a standard skeletal kinematic tree with $K=24$ joints. For simplicity, we include the root orientation as part of the pose parameters for the root joint in our formulation. The shape parameters $\boldsymbol{\beta} \in \mathbb{R}^{10}$ indicate the body shape configurations, such as height. Given the pose and shape parameters, the model deforms accordingly and generates a triangulated mesh comprising $N=6890$ vertices as $\mathcal{M}(\boldsymbol{\theta}, \boldsymbol{\beta}) \in \mathbb{R}^{N \times 3}$. The deformation process $\mathcal{M}(\boldsymbol{\theta}, \boldsymbol{\beta})$ is differentiable with respect to the pose $\boldsymbol{\theta}$ and shape $\boldsymbol{\beta}$ parameters.
Once the final mesh is obtained, sparse 3D keypoints can be mapped from vertices through a pretrained linear regressor.
Other models such as SMPL-X~\cite{SMPL-X:2019} extends SMPL~\cite{loper2015smpl} model and constructs a comprehensive model, wherein the body, face, and hands are modeled jointly. 
In addition to SMPL-based linear models, alternative modeling approaches have been explored, such as GHUM~\cite{xu2020ghum} and STAR~\cite{STAR:2020}. \revision{To ensure conciseness, we employ the shorthand term ``Rot.'' in the tables below to encompass both joint-based 3D rotations and their applications in statistical human models (\eg SMPL), without delving into intricate differentiation between the two.}

\subsubsection{Motion Data Collection}
\label{section:motion_data_collection}
\begin{figure}
    \centering
    \includegraphics[width=\linewidth]{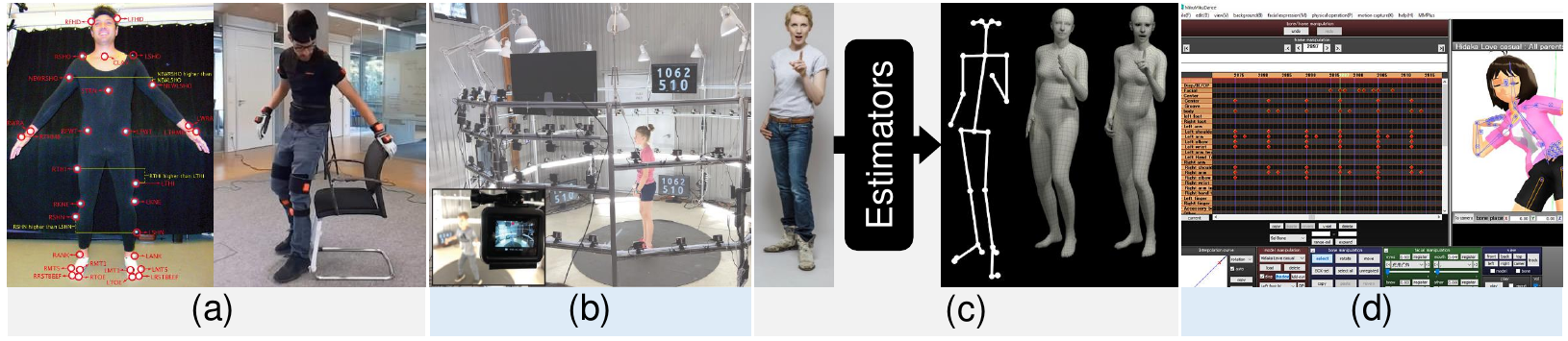}
    \caption{Human motion data collection methods. (a) Examples of marker-based motion capture setup where (left) optical markers \cite{cmuWEB} or (right) IMUs \cite{zhang2022couch} are attached to the subject's body surface. (b) Example of the markerless multiview motion capture system \cite{HUMBI}. (c) Pseudo-labeling pipeline involves using pose or mesh estimators to generate pseudo labels \cite{pavlakos2019expressive}. (d) Example interface for manual collection using MikuMikuDance (MMD) resources.}
    \vspace{-4mm}
    \label{fig:mocap}
\end{figure}

There are four main approaches to collecting human motion data: (i) \emph{marker-based} motion capture, (ii) \emph{markerless} motion capture, (iii) \emph{pseudo-labeling}, and (iv) \emph{manual annotation}.

\noindent\textbf{Marker-based} motion capture involves placing small reflective markers or Inertial Measurement Units (IMUs) at specific locations in the subject's body and then tracking the movement of these markers in a 3D space. See Figure \ref{fig:mocap} (a) for illustration. This data can then be used to obtain 3D keypoints by applying forward kinematics \cite{h36m_pami} or a parametric body mesh such as SMPL \cite{loper2015smpl} with the help of MoSh \cite{MoSh_lopermahmoodetal2014}. Optical markers provide more accurate data than IMUs, but are less portable and are typically used in indoor environments, while IMUs can be used in outdoor settings.

\noindent\textbf{Markerless} motion capture solutions track the movement of the subject's body without the need for markers from one or multiple cameras and use computer vision algorithms (\eg, ~\cite{iskakov2019learnable,voxelpose,fastervoxelpose}) to get the 3D motion by exploiting multi-view geometry, as shown in Figure \ref{fig:mocap} (b). Multiple RGB or RGB-D cameras will be set up and synchronized during the capture process. This solution is less accurate than marker-based motion capture, but is more convenient and can be used in a wider range of settings.

\noindent\textbf{Pseudo-labeling} of human motion is primarily intended for in-the-wild captured monocular RGB images or videos. This involves predicting 2D or 3D human keypoints with existing human pose estimators such as OpenPose \cite{cao2018openpose} and VideoPose3D \cite{pavllo20193d}, or fits body model to image evidence to generate pseudo 3D mesh labels, \eg, by using SMPLify-X \cite{pavlakos2019expressive}. See Figure \ref{fig:mocap} (c). However, pseudo-labels tend to have more errors compared to motion capture systems.

\noindent\textbf{Manual annotation} involves designing human motion with an animation engine manually, typically using a team of skilled artists. Figure \ref{fig:mocap} (d) shows an example engine interface of MikuMikuDance (MMD). While this approach can produce high-quality animations, it is expensive, time-consuming, and not scalable.

\subsection{Motion Generation Methods}

\begin{figure}[t]
    \centering
    \includegraphics[width=\linewidth]{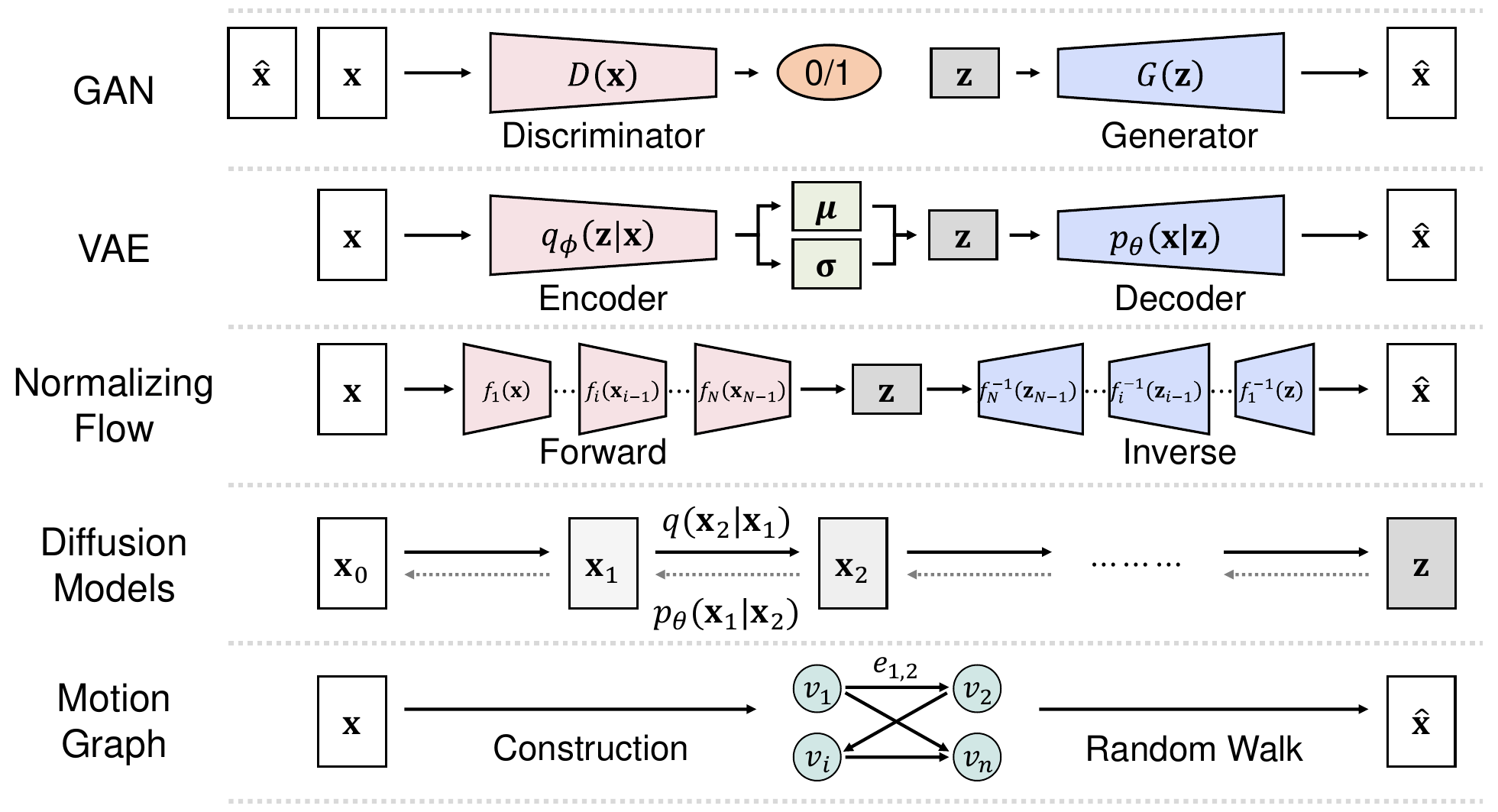}
    \caption{
        An overview of different generative models.
    }
        \vspace{-3mm}
    \label{fig:generative-models}
\end{figure}

We roughly classify human motion generation methods into two classes. The first class of methods is based on \emph{regression models} to predict human motion using features encoded from input conditions. They fall into the supervised learning paradigm and aim to establish a direct mapping from input conditions to target motions. The other class of methods base on \emph{generative models}. They focus on modeling the underlying distribution of motion (or joint distribution with conditions) in an unsupervised manner. Typical deep generative models include Generative Adversarial Networks (GANs),  Variational Autoencoders (VAEs), Normalizing Flows, and Denoising Diffusion Probabilistic Models (DDPMs). In addition to the general generative models, a task-specific model, motion graph, has also been widely used especially in the field of computer graphics and animation. 
Figure~\ref{fig:generative-models} shows an overview of different generative models. 
In the following, we will briefly go over commonly used generative models in motion generation.

\noindent\textbf{Generative Adversarial Networks.}
\revision{GANs~\cite{gan} are a class of generative models composed of two neural networks: the generator $G$ and the discriminator $D$. The generator produces synthetic data from a noise vector $\z$ to deceive the discriminator. Conversely, the discriminator tries to differentiate between real data and synthetic data produced by the generator. This dynamic between the generator and the discriminator can be viewed as a zero-sum or min-max game. The loss function representing their interaction can be formulated as:}
\begin{align}
    \Loss_D & = - \E_{\x\sim p_{\x}} [\log(D(\x))] - \E_{\z\sim p_{\z}}[\log(1-D(G(\z)))], \label{eqa:loss_d} \\
    \Loss_G & = - \E_{\z\sim p_{\z}}[\log(D(G(\z)))]. \label{eqa:loss_g}
\end{align}
\revision{With the rise of deep learning, various deep learning-based GANs have been proposed. Models such as DCGAN \cite{radford2015unsupervised}, PGGAN\cite{karras2017progressive}, and StyleGAN\cite{Karras_2019_CVPR, Karras_2020_CVPR} have demonstrated remarkable achievements and potential. These advancements in GANs have contributed significantly to the field of generative models, especially in the generation of synthetic data. However, GANs face several challenges, including training instability, convergence problems, and mode collapse.}

\noindent\textbf{Variational Autoencoders.}
\revision{VAEs~\cite{vae} are notable generative models that provide robust solution for data representation. They address the challanges of interactable likelihood by using feed-forward model, denoted as $q_{\phi}(\z|\x)$, to approximate the intractable posterior. The primary optimization goal is to minimize the KL divergence between this approximation and the original posterior. VAEs adopt the Evidence Lower Bound (ELBO) as loss function:
}
\begin{align}
    \text{ELBO} = \E_{\z\sim q_{\phi(\z|\x)}}\mathrm{log}(p_\theta(\x|\z)) - D_{KL}(q_{\phi}(\z|\x) || p_\theta(\z)) 
\end{align}
\revision{VAEs efficiently generate and infer new samples due to the feed-forward mode of $q_{\phi}(\z|\x)$. Additionally, the reparameterization trick enables differentiable sample generation and the utilization of a reconstruction-based loss function, ultimately enhancing training efficiency and stability. These advantages have led to the widespread adoption of VAEs variants, such as CVAE~\cite{NIPS2015_8d55a249}, LVAE~\cite{NIPS2016_6ae07dcb}, and VQ-VAE~\cite{oord2018neural}, in various fields and drive advances in generative models. However, VAEs are subject to the risk of posterior collapse and may produce less sharp samples compared to GANs.}

\noindent\textbf{Normalizing Flows.}
GANs and VAEs implicitly learn the probability density of data. They can hardly calculate the exact likelihood. In contrast, Normalizing Flows~\cite{normalizingflows} is a class of generative models that explicitly learn the data distribution $p\left(\x\right)$ and allows for tractable probability density estimation. These models employ a series of invertible transformations $\{f_{i}\}_{1:N}$ to map a simple prior distribution $p(\z_0)$ (e.g., a standard Gaussian) to a complex data distribution $p(\x)$:
\begin{align}
    \z_{i} &= f_{i}\left( \z_{i-1} \right) \\
    \x = \z_N &= f_N \circ f_{N-1} \circ \dots \circ f_1 (\mathbf{z}_0)
\end{align}
The density of the target distribution can be obtained by applying the change of variables theorem:
\begin{align}
\log p(\z_i) &= \log p(\z_{i-1}) - \log \left\vert \det \dfrac{d f_i}{d\z_{i-1}} \right\vert \\
\log p(\x) &= \log p(\z_0) - \sum_{i=1}^K \log \left\vert \det \dfrac{d f_i}{d\z_{i-1}}\right\vert
\end{align}
where $\det$ is short for the determinant of a square matrix. Normalizing Flows can be typically trained by maximizing the log-likelihood of the observed data. Owing to the invertible transformation, Normalizing Flows offer flexibility, exact likelihood computation, and easy data sampling. However, they require a large number of transformations to model complex distributions and can be computationally expensive and difficult to train.

\noindent\textbf{Diffusion Models.}
Diffusion models~\cite{DDPM,weng2021diffusion,song2020score} define a forward diffusion process that gradually adds a small amount of Gaussian noise to the input data $\x_{0}$ in $T$ steps, producing a series of noisy samples $\{\x_{t}\}_{1:T}$. Noise is scheduled by $\{\beta_{t}\}_{1:T}$.
\begin{align}
    q(\x_t|\x_{t-1}) &= \mathcal{N}(\x_t; \sqrt{1-\beta_t}\x_{t-1}, \beta_t \mathbf{I}), \\
    q(\x_{1:T}|\x_0) &= \prod_{t=1}^{T} q(\x_t|\x_{t-1}).
\end{align}
As $T \rightarrow \infty$, $\x_T$ is actually a Gaussian distribution. If we know the reverse transition $q(\x_{t-1}|\x_t)$, then we can sample from a Gaussian prior $\x_t \sim \mathcal{N}(0, \mathbf{I})$ and run the diffusion process in reverse to get a sample from the real data distribution $p(\x_0)$.
\revision{However, since $q(\x_{t-1}|\x_t)$ depends on the entire dataset and is hard to estimate, we train a neural network $p_{\theta}$ to match the posterior $q(\x_{t-1}|\x_t, \x_0)$, a tractable Gaussian, instead of $q(\x_{t-1}|\x_t)$:}
\begin{align}
    p_\theta(\mathbf{x}_{t-1} \vert \mathbf{x}_t) &= \mathcal{N}(\mathbf{x}_{t-1}; \boldsymbol{\mu}_\theta(\mathbf{x}_t, t), \boldsymbol{\Sigma}_\theta(\mathbf{x}_t, t))
\end{align}
$p_{\theta}$ is learned by optimizing the ELBO like VAE. In practice, diffusion models are able to produce high-quality samples and can benefit from stable training. However, it relies on a long Markov chain of reverse diffusion steps to generate samples, so it can be computationally expensive and slower than GANs and VAEs.

\noindent\textbf{Motion Graph.} Motion graph~\cite{arikan2002interactive, lee2002interactive, motiongraph2002} can be represented mathematically as a directed graph $G = \langle V, E \rangle$ where $V$ denotes the set of nodes or vertices, and $E$ denotes the set of directed edges or transitions. Each node $v \in V$ represents a pose or keyframe, and each directed edge $e \in E$ connects two vertices $(v_1, v_2)$ and represents a feasible transition between the corresponding poses. Motion graphs are first constructed based on a collection of motion clips. To ensure smooth transitions, the algorithm identifies compatible poses within the motion clips and connects them with edges, forming a graph that can be traversed to generate new motion sequences. After constructing the motion graph, a random walk $W = (v_1, v_2, \dots, v_n)$ can be performed on the graph, starting from an initial node and following the directed edges. The output motion sequence is a concatenation of the poses corresponding to the traversed nodes, ensuring smooth transitions between consecutive poses. Meanwhile, further constraints can be incorporated as optimization objectives~\cite{shiratori2006dancing, choreomaster2021}. This process effectively creates new motion sequences that were not explicitly present in the original dataset, but are consistent with the overall characteristics of the data.

\begin{table*}[!h]

\caption{Representative works of human motion generation. ``Kpts.'' and ``Rot.'' denotes keypoints and 3D rotations, respectively.
$^{\dagger}$ denotes optional condition. 
}

\center
\small
\setlength{\tabcolsep}{5pt}
\resizebox{\linewidth}{!}{
\begin{tabular}{l | c | c | c | c | c}
\toprule
\textbf{Method} & \textbf{Venue} & \textbf{Representation} & \textbf{Model} & \textbf{Condition} & \textbf{Dataset} \\
\toprule

Action2Motion~\cite{guo2020action2motion} & MM 2020 & Rot. & VAE & Action class & \cite{guo2020action2motion, liu2020ntu}\\

SA-GCN~\cite{yu2020structure} & ECCV 2020 & Kpts. (2D, 3D) & GAN & Action class & \cite{liu2020ntu, ionescu2013human3}\\

ACTOR~\cite{petrovich21actor} & ICCV 2021 & Rot. & VAE & Action class & \cite{guo2020action2motion, liu2020ntu, ji2019large}\\

Kinetic-GAN~\cite{degardin2022generative} & WACV 2022 & Kpts. (3D) & GAN & Action class & \cite{liu2020ntu, ionescu2013human3, shahroudy2016ntu}\\

ODMO~\cite{lu2022action} & MM 2022 & Kpts. (3D) & VAE & Action class, Trajectory$^{\dagger}$ & \cite{guo2020action2motion, ji2019large}\\

PoseGPT ~\cite{posegpt} & ECCV 2022 & Rot. & VAE & Action class, Duration, Past motion$^{\dagger}$ & \cite{guo2020action2motion, punnakkal2021babel, taheri2020grab}\\

Cervantes ~\etal ~\cite{cervantes2022implicit} & ECCV 2022 & Rot. & Regression & Action class & \cite{guo2020action2motion, liu2020ntu, ji2019large}\\

MultiAct~\cite{Lee2023MultiAct} & AAAI 2023 & Kpts. (3D) / Rot. & VAE & Action class, Past motion$^{\dagger}$ & \cite{punnakkal2021babel}\\

MDM~\cite{tevet2022human} & ICLR 2023  & Kpts. (3D) / Rot. & Diffusion & Action class / Text & \cite{Guo_2022_CVPR, guo2020action2motion, ji2019large, Plappert2016}\\

MLD~\cite{chen2022mld} & CVPR 2023 & Kpts. (3D) / Rot. & Diffusion, VAE & Action class / Text & \cite{Guo_2022_CVPR, guo2020action2motion, ji2019large, Plappert2016}\\

\hdashline 

Text2Action~\cite{ahn2018text2action} & ICRA 2018 & Kpts. (3D) & GAN & Text & \cite{xu2016msr}\\

JL2P~\cite{8885540} & 3DV 2019 & Kpts. (3D) & Regression & Text & \cite{Plappert2016}\\

Ghosh ~\etal ~\cite{Ghosh_2021_ICCV} & ICCV 2021 & Kpts. (3D) & GAN & Text & \cite{Plappert2016}\\

Guo ~\etal ~\cite{Guo_2022_CVPR} & CVPR 2022 & Kpts. (3D) / Rot. & VAE & Text, POS & \cite{Guo_2022_CVPR, Plappert2016} \\

AvatarCLIP~\cite{hong2022avatarclip} & TOG 2022  & Rot. & VAE & Text & \cite{AMASS:2019}\\

MotionCLIP~\cite{tevet2022motionclip} & ECCV 2022 & Rot. & Regression & Text & \cite{punnakkal2021babel}\\

TEMOS~\cite{petrovich22temos} & ECCV 2022 & Kpts. (3D) / Rot. & VAE & Text & \cite{Plappert2016}\\

TM2T~\cite{chuan2022tm2t} & ECCV 2022 & Kpts. (3D) / Rot. & VAE & Text & \cite{Guo_2022_CVPR, Plappert2016}\\

TEACH~\cite{TEACH:3DV:2022} & 3DV 2022 & Rot. & VAE & Text, Past motion$^{\dagger}$ & \cite{punnakkal2021babel}\\

FLAME~\cite{https://doi.org/10.48550/arxiv.2209.00349} & AAAI 2023 & Kpts. (3D) / Rot. & Diffusion & Text & \cite{Guo_2022_CVPR, punnakkal2021babel, Plappert2016}\\

T2M-GPT~\cite{zhang2023generating} & CVPR 2023 & Kpts. (3D) / Rot. & VAE & Text & \cite{Guo_2022_CVPR, Plappert2016}\\

OOHMG~\cite{lin2023comes} & CVPR 2023 & Rot. & VAE & Text & \cite{AMASS:2019, punnakkal2021babel}\\

UDE~\cite{ude2022} & CVPR 2023 & Rot. & Diffusion, VAE & Text / Music & \cite{Guo_2022_CVPR, li2021learn}\\

MoFusion~\cite{dabral2022mofusion} & CVPR 2023 & Kpts. (3D) & Diffusion & Text / Music & \cite{Guo_2022_CVPR, punnakkal2021babel, li2021learn}\\

\midrule

Dance with Melody~\cite{tang2018dance} & MM 2018 & Kpts. (3D) & Regression & Music & ~\cite{tang2018dance} \\

Dancing to Music~\cite{NEURIPS2019_7ca57a9f} &  NeurIPS 2019 & Kpts. (2D) & GAN & Music & \cite{NEURIPS2019_7ca57a9f}  \\

Dance Revolution~\cite{huang2021} &  ICLR 2021 & Kpts. (2D) & Regression & Music & \cite{huang2021} \\

AI Choreographer~\cite{li2021learn} &  ICCV 2021 & Rot. & Regression & Music, Past motion & \cite{li2021learn}  \\

Transflower~\cite{Valle-Pérez2021Transflower} & TOG 2021 & Kpts. (3D) & Normalizing Flow & Music, Past motion & \cite{li2021learn, Valle-Pérez2021Transflower, alemi2017groovenet, holzapfel2020diversity} \\

ChoreoMaster~\cite{choreomaster2021} &  TOG 2021 & Rot. & Motion Graph & Music & \cite{choreomaster2021}  \\

DanceFormer~\cite{DBLP:conf/aaai/LiZZS22} & AAAI 2022 & Rot. & GAN & Music & \cite{DBLP:conf/aaai/LiZZS22}  \\

Bailando~\cite{siyao2022bailando} & CVPR 2022 & Kpts. (3D) & VAE & Music, Past motion & \cite{li2021learn} \\

MNET~\cite{Kim_2022_CVPR} & CVPR 2022 & Rot. & GAN & Music, Past motion, Style code & \cite{li2021learn} \\

PC-Dance~\cite{PC-Dance} & MM 2022 & Rot. & Motion Graph & Music, Anchor pose$^{\dagger}$ & \cite{PC-Dance} \\

ChoreoGraph~\cite{ChoreoGraph2022} & MM 2022 & Kpts. (3D) & Motion Graph & Music & \cite{li2021learn} \\

GroupDancer~\cite{10.1145/3503161.3548090} & MM 2022 & Rot. & Regression & Music & \cite{10.1145/3503161.3548090} \\

Sun ~\etal ~\cite{sun2022you} & NeurIPS 2022 & Rot. & VAE & Music, Past motion & \cite{li2021learn} \\

Aristidou~\etal~\cite{9745335} & TVCG 2022 & Rot. & Regression & Music & \cite{9745335, li2021learn} \\

EDGE~\cite{tseng2022edge} & CVPR 2023 & Rot. & Diffusion & Music & \cite{li2021learn} \\

GDanceR~\cite{aiozGdance} & CVPR 2023 & Rot. & Regression & Music & \cite{aiozGdance} \\

\hdashline 

Ginosar~\etal~\cite{ginosar2019gestures} & CVPR 2019 & Kpts. (2D) & GAN & Speech & \cite{ginosar2019gestures} \\

Aud2Repr2Pose~\cite{Aud2Repr2Pose} & IVA 2019 & Kpts. (3D) & Regression & Speech & \cite{takeuchi2017creating} \\

Mix-StAGE~\cite{ahuja2020style} & ECCV 2020 & Kpts. (2D) & GAN & Speech, Style code & \cite{ginosar2019gestures, ahuja2020style} \\

StyleGestures~\cite{alexanderson2020style} & CGF 2020 & Rot. & Normalizing Flow & Speech, Past motion & \cite{Trinity2018} \\

Trimodal Context~\cite{Yoon2020Speech} & TOG 2020 & Kpts. (3D) & GAN & Speech, Text, Speaker, Past motion & \cite{yoon2019robots, Yoon2020Speech} \\

Habibie ~\etal ~\cite{habibie2021learning} & IVA 2021 & Kpts. (3D) & GAN & Speech & \cite{ginosar2019gestures,  habibie2021learning} \\

S2AG~\cite{bhattacharya2021speech2affectivegestures} & MM 2021 & Kpts. (3D) & GAN & Speech, Text, Speaker, Past motion & \cite{yoon2019robots, 10.1145/3397481.3450692} \\

Qian~\etal~\cite{qian2021speech} & ICCV 2021 & Kpts. (2D) & VAE & Speech, Template vector & \cite{ginosar2019gestures} \\

Audio2Gestures~\cite{li2021audio2gestures} & ICCV 2021 & Kpts. (2D) / Rot. & VAE & Speech & \cite{Trinity2018, ginosar2019gestures} \\
        
HA2G~\cite{Liu_2022_CVPR} & CVPR 2022 & Kpts. (3D) & GAN & Speech, Text, Speaker, Past motion & \cite{yoon2019robots, Yoon2020Speech}\\

DisCo~\cite{disco2022} & MM 2022 & Rot. & GAN & Speech, Past motion & \cite{Trinity2018, ginosar2019gestures}\\

CaMN~\cite{liu2022beat} & ECCV 2022 & Rot. & GAN & Speech, Text, Speaker, Expressions, Emotions, Semantics & \cite{ginosar2019gestures}, \cite{liu2022beat}\\

Habibie~\etal~\cite{10.1145/3528233.3530750} & TOG 2022 & Kpts. (3D) & GAN & Speech, Start pose, Control$^{\dagger}$ & \cite{ginosar2019gestures}, ~\cite{habibie2021learning} \\

Rhythmic Gesticulator ~\cite{10.1145/3550454.3555435} & TOG 2022 & Rot. & VAE & Speech, Text, Speaker, Past motion & \cite{Trinity2018}, ~\cite{yoon2019robots}, ~\cite{10.1145/3550454.3555435} \\

DiffGesture~\cite{zhu2023taming} & CVPR 2023 & Kpts. (3D) & Diffusion & Speech & \cite{yoon2019robots, Yoon2020Speech}\\

TalkSHOW~\cite{yi2022generating} & CVPR 2023 & Rot. & VAE & Speech, Speaker & \cite{yi2022generating}\\

QPGesture~\cite{yang2023QPGesture} & CVPR 2023 & Rot. & VAE & Speech, Text, Anchor pose, Control$^\dagger$ & \cite{liu2022beat}\\

LDA~\cite{alexanderson2022listen} & TOG 2023 & Rot. & Diffusion & Speech / Music / Path, Style code$^{\dagger}$ & \cite{10.1145/3397481.3450692, ghorbani2022zeroeggs, alemi2017groovenet, Valle-Pérez2021Transflower, 10.1145/3522618}\\

GestureDiffuCLIP~\cite{Ao2023GestureDiffuCLIP} & TOG 2023 & Rot. & Diffusion, VAE & Speech, Text, Style prompt & \cite{ghorbani2022zeroeggs, liu2022beat}\\

\midrule

Corona~\etal~\cite{corona2020context} & CVPR 2020 & Kpts. (3D) & Regression & Scene (object), Past motion & \cite{KIT_Dataset, cmuWEB} \\
Cao~\etal~\cite{cao2020long} & ECCV 2020 & Kpts. (3D) & VAE & Scene (image), Past motion & \cite{hassan2019resolving, cao2020long} \\

Wang~\etal~\cite{wang2021synthesizing} & CVPR 2021 & Rot. & VAE & Scene (mesh), Start pose, End pose, Sub-goal & \cite{hassan2019resolving, chang2017matterport3d} \\
Wang~\etal~\cite{wang2021scene} & CVPR 2021 & Kpts. (3D) & GAN & Scene (image), Start pose & \cite{cao2020long, hassan2019resolving} \\
HuMoR~\cite{rempe2021humor} & ICCV 2021 & Rot. & VAE & Scene (ground), Past motion & \cite{AMASS:2019}, \cite{hassan2019resolving}, \cite{iMapper} \\
SAMP~\cite{hassan2021stochastic} & ICCV 2021 & Rot. & VAE & Scene (interactive object), Action class & \cite{hassan2021stochastic} \\
Wang~\etal~\cite{wang2022towards} & CVPR 2022 & Rot. & VAE & Scene (mesh), Action class & \cite{hassan2019resolving, chang2017matterport3d} \\
GAMMA~\cite{zhang2022wanderings} & CVPR 2022 & Kpts. (3D) & RL & Scene (goal)  & \cite{AMASS:2019} \\
GOAL~\cite{taheri2022goal} & CVPR 2022 & Rot. & VAE & Scene (object), Start pose & \cite{taheri2020grab} \\
SAGA~\cite{wu2022saga} & ECCV 2022 & Kpts. (3D) & VAE & Scene (object), Start pose & \cite{taheri2020grab} \\
Couch~\cite{zhang2022couch} & ECCV 2022 & Rot. & Regression & Scene (chair, contact), Start pose & \cite{zhang2022couch} \\
Mao~\etal~\cite{mao2022contact} & NeurIPS 2022 & Kpts. (3D) & Regression & Scene (point cloud), Past motion & \cite{cao2020long}, \cite{hassan2019resolving} \\
HUMANISE~\cite{wang2022humanise} & NeurIPS 2022 & Rot. & VAE & Scene (point cloud), Text & \cite{wang2022humanise} \\
IMoS~\cite{ghosh2022imos} & EUROGRAPHICS 2023 & Rot. & VAE & Scene (object), Text & \cite{taheri2020grab} \\  
SceneDiffuser~\cite{huang2023diffusion} & CVPR 2023 & Rot. & Diffusion & Scene (point cloud, goal) & \cite{hassan2019resolving} \\
CIRCLE~\cite{araujo2023circle} & CVPR 2023 & Rot. & Regression & Scene (point cloud, goal), Start pose & \cite{araujo2023circle} \\
\bottomrule 
\end{tabular}
}

\label{tab:methods}
\end{table*}

\section{Text-Conditioned Motion Generation}
\label{sec:text2motion}
Text possesses a remarkable ability to convey various types of actions, velocities, directions, and destinations, either explicitly or implicitly. This characteristic makes the text an appealing condition for generating human motion. This section aims to elucidate the topic of text-conditioned human motion generation tasks (see Table \ref{tab:methods} top block), which can be primarily divided into two categories: \emph{action-to-motion} and \emph{text-to-motion}. 

\subsection{Action to Motion}
\revision{The action-to-motion task is focused on generating human motion sequences based on specific action categories, such as `Walk', `Kick', or `Throw'. These actions are often represented using techniques like one-hot encoding, which simplifies the motion generation process. Compared to text-to-motion tasks which deal with the complexities of natural language processing, this representation provides a more straightforward task due to limited and well-defined action classes.
}

Yu~\etal~\cite{yu2020structure} introduce SA-GAN, which leverages a self-attention-based graph convolutional network (GCN) with GAN architecture. They also propose to enhance the generative capabilities through the use of two discriminators - one frame-based and the other sequence-based. In a similar vein, Kinetic-GAN~\cite{degardin2022generative} combines the strengths of GAN and GCN, and further utilizes latent space disentanglement and stochastic variations to generate high-quality and diverse human motions.
Guo~\etal~\cite{guo2020action2motion} introduce Action2Motion, a per-frame VAE architecture based on Gated Recurrent Units (GRU) to generate motion sequences. Similarly, ACTOR~\cite{petrovich21actor} employs a sequence-level CVAE model that uses transformers as a backbone for generating motion sequences non-autoregressively. This non-autoregressive approach allows for the one-shot generation of motion sequences. ODMO~\cite{lu2022action} adopts a novel strategy of applying contrastive learning within a low-dimensional latent space, thus generating hierarchical embeddings of motion sequences. The model initially creates motion trajectories before generating the motion sequences, thus benefiting trajectory control. Furthermore, PoseGPT~\cite{posegpt} utilizes an auto-regressive transformer to encode human motion into quantized latent representations, subsequently employing a GPT-like model for next motion index predictions within this discrete space. Cervantes~\etal~\cite{cervantes2022implicit} introduce a method that uses implicit neural representation (INR) and a fitted conditional Gaussian mixed model (GMM). This method controls the length and action classes of the sequence by extracting representations from the variational distributions of each training sequence. In addition, MDM~\cite{tevet2022human} utilizes a diffusion model to predict samples at each diffusion step, rather than just noise. MLD~\cite{chen2022mld} draws inspiration from the Latent Diffusion Model (LDM)~\cite{rombach2021highresolution} to employ latent-level diffusion and VAE for motion generation.

While these methods have greatly advanced the field of action-to-motion, they primarily excel at generating single-action motions. The transition to generating complex sequences that involve multiple actions remains a challenge and often requires additional post-processing to connect disparate actions. To this end, a recent work, MultiAct~\cite{Lee2023MultiAct}, leverages past motion to recurrently generate long-term multi-action 3D human motion and proposes a novel face-front canonicalization methodology to ensure the local coordinate system shares the ground geometry in each recurrent step.

\subsection{Text to Motion}
\revision{The text-to-motion task aims to generate human motion sequences from natural language descriptions, leveraging the vast expressive power of language. In contrast to action-to-motion, which utilizes limited predefined labels, text-to-motion has the potential to produce a wider variety of motions based on diverse textual descriptions. Nonetheless, the challenge lies in accurately converting the intricacies of text into corresponding movements, requiring a profound understanding of both linguistic nuances and physical motion dynamics.
}

Text2Action~\cite{ahn2018text2action} first leverages GAN to generate a variety of motions from a given natural language description. Some other methods have explored the potential of learning a joint embedding of text and motion. For instance, JL2P~\cite{8885540} uses a GRU-based text encoder and motion encoder-decoder to map the text into a corresponding human motion. Ghosh~\etal~\cite{Ghosh_2021_ICCV} further develop a two-stream encoder-decoder model for co-embedding text and body movements, while also employing a GAN structure to generate more natural motions. Guo~\etal~\cite{Guo_2022_CVPR} propose a VAE-based approach that utilizes a length estimation module and a word-level attention module at each frame to produce diverse multi-length motions. Additionally, TEMOS~\cite{petrovich22temos} learns the joint distribution of the motion and the text through a VAE with Transformer layers, enabling the generation of varied motion sequences. TEACH~\cite{TEACH:3DV:2022} further employs past motions as supplementary inputs to the encoder module, which enables the generation of more natural and coherent motion sequences, especially when dealing with several sequences of text inputs.

While the above methods pay attention to generating motion based on a given dataset, they may encounter inherent challenges when it comes to zero-shot generation. To address this challenge, MotionCLIP~\cite{tevet2022motionclip} utilizes a Transformer-based autoencoder and aligns the motion latent space with the text and image space of a pre-trained vision-language model CLIP~\cite{radford2021learning} to enhance the zero-shot generation ability. AvatarCLIP~\cite{hong2022avatarclip} also employs CLIP~\cite{radford2021learning} and a reference-based motion synthesis method to generate diverse animations from natural language descriptions. Furthermore, OOHMG~\cite{lin2023comes} uses a text-pose generator to obtain text-consistent poses, which are then fed as masked prompts into a pre-trained generator. This allows for efficient full-motion reconstruction, eliminating the need for paired data or online optimization. It is worth noting that while these methods utilize text as input, they only employ short text that primarily consists of the action class name.

In recent years, there has been a growing interest in VQ-VAE and Diffusion models inspired by their remarkable success in the field of text-to-image generation. For instance, TM2T~\cite{chuan2022tm2t} exploits VQ-VAE to train the text-to-motion and motion-to-text modules in tandem. Similarly, T2M-GPT~\cite{zhang2023generating} applies a GPT-like transformer architecture for motion sequence generation, combining VQ-VAE with an Exponential Moving Average (EMA) and code reset strategy. FLAME~\cite{https://doi.org/10.48550/arxiv.2209.00349} proposes to concatenate the motion-length token, language pooler token, time-step token, and motion embeddings, which are then utilized by the diffusion model to generate variable-length and diverse motions. MDM~\cite{tevet2022human} and MLD~\cite{chen2022mld}, already introduced in the action-to-motion section, also apply the aforementioned methods for text-to-motion generation. Several works further explore motion generation from various conditions. For instance, MoFusion~\cite{dabral2022mofusion} utilizes a diffusion model with 1D U-Net style Transformer module to generate human motion sequences from either natural language or audio input. Furthermore, Zhou et al.~\cite{ude2022} introduce UDE, a framework that discretizes motion sequences into latent codes, maps conditions into a unified space, predicts the quantized codes using a GPT-style transformer, and generates motions via a diffusion model.

\section{Audio-Conditioned Motion Generation}

\label{sec:audio2motion}

In addition to textual descriptions, human motion generation from audio signals has also been explored. Unlike text, audio signals typically do not provide explicit depictions of the corresponding human motions, resulting in a higher degree of freedom for the generative task. Meanwhile, the generated human motion should be harmonious with the audio in terms of both high-level semantics and low-level rhythms. In this section, we mainly discuss two subtasks of increasing attention: \emph{music-to-dance} and \emph{speech-to-gesture}. \revision{The audio conditions can be represented by raw audio waveform, spectrogram, and mel-frequency cepstrum coefficients (MFCC). To enhance controllability, some works incorporate additional conditions such as style code or textual transcripts.}
Please refer to Table \ref{tab:methods} middle block for a summary of the methods.

\subsection{Music to Dance}

The music-to-dance generation task aims to generate corresponding dance moves given an input music sequence. One straightforward idea is to approach the problem using fully-supervised \emph{regression models}, similar to sequence-to-sequence translation. For instance, Tang~\etal~\cite{tang2018dance} employ an LSTM autoencoder to extract acoustic features and translate them to motion features. AI Choreographer~\cite{li2021learn} utilizes a full-attention cross-modal Transformer (FACT) and predicts $N$ future motion frames in an auto-regressive manner. GroupDancer~\cite{10.1145/3503161.3548090} proposes an additional dancer collaboration stage to select active dancers to generate multi-person dance. GDancer~\cite{aiozGdance} introduces global-local motion representations to ensure both local coherency and global consistency.
The above methods adopt a fully supervised learning perspective to minimize the distance between the predicted and ground truth motions. Nevertheless, for a given music sequence, there exists a wide variety of plausible dancing motions. Simple reconstruction supervision does not adequately address this one-to-many mapping relationship.

From a generative perspective, GAN-based methods ~\cite{NEURIPS2019_7ca57a9f, DBLP:conf/aaai/LiZZS22} apply adversarial learning to regularize the distance between generated and real motion data manifolds. MNET~\cite{Kim_2022_CVPR} additionally incorporates a music style code for the generator and designs a multi-task discriminator to perform per-style classification. Transflower~\cite{Valle-Pérez2021Transflower} utilizes normalizing flow to express the complex probability distributions over valid motions. Bailando~\cite{siyao2022bailando} first quantizes 3D motions using a VQ-VAE codebook, then leverages an actor-critic Generative Pretrained Transformer (GPT) to compose coherent sequences from the learned latent codes. EDGE~\cite{tseng2022edge} builds upon the diffusion model and formulates the task as a motion denoising problem conditioned on music. Another class of approaches is based on the classical motion graph framework, which casts motion generation as solving for an optimal path in a pre-constructed graph. ChoreoMaster~\cite{choreomaster2021} proposes to learn a shared embedding space of music and dance, then integrate the learned embeddings and expert knowledge into the graph-based motion synthesis framework. PC-Dance~\cite{PC-Dance} further achieves controllable motion generation by incorporating anchor poses as additional inputs. ChoreoGraph~\cite{ChoreoGraph2022} utilizes motion segment warping to address rhythm alignment issues, reducing motion nodes in the graph and computation cost.

While most methods utilize short music-dance clips for training, an important user demand is to generate perpetual dance for an entire song. However, long-sequence generation tends to incur error accumulation issues that result in freezing motions.
To overcome this challenge, Huang~\etal~\cite{huang2021} propose a curriculum learning approach that progressively transitions from a teacher-forcing scheme to an autoregressive scheme as training advances. Sun~\etal~\cite{sun2022you} employ a VQ-VAE to learn a low-dimensional manifold, which effectively denoises the motion sequences. They also develop a past-future motion dynamics bank to provide explicit priors about future motions. Aristidou~\etal~\cite{9745335} address the problem from three levels, including pose, motif, and choreography, to generate long dances that maintain a genre-specific global structure.

\subsection{Speech to Gesture}

The speech-to-gesture generation (or co-speech gesture synthesis) task aims to generate a sequence of human gestures based on input speech audio and, in some cases, text transcripts. Co-speech gestures play a crucial role in non-verbal communication, conveying the speaker's information and emotions, fostering intimacy, and enhancing trustworthiness~\cite{van1998persona}. Existing research works for this task generally focus on upper-body motion, as lower-body movement tends to be static.

Some studies generate speech gestures from text transcripts~\cite{ishi2018speech, yoon2019robots, bhattacharya2021text2gestures}. A greater number of research works focus on speech audio conditions. For instance, Ginosar~\etal ~\cite{ginosar2019gestures} collect a speech video dataset of person-specific gestures and train a generative model with adversarial loss. Aud2Repr2Pose~\cite{Aud2Repr2Pose} first constructs a motion autoencoder and then trains a speech encoder to map the speech audio to motion representations. StyleGestures~\cite{alexanderson2020style} adapts MoGlow~\cite{henter2020moglow} and further exerts directorial control over the styles of the generated motions. Recognizing that speech cannot fully determine gestures, Qian~\etal~\cite{qian2021speech} propose to learn a set of gesture template vectors to model the general appearance of generated gestures. Audio2Gestures~\cite{li2021audio2gestures} disentangles motion representation into audio-motion shared and motion-specific information to reflect the one-to-many mapping between audio and motion. 
Habibie~\etal~\cite{habibie2021learning} apply an audio encoder and three separate decoders for face, body, and hand respectively. DisCo~\cite{disco2022} first clusters the motion sequences into content and rhythm segments, then trains on the content-balanced data distribution. Habibie~\etal~\cite{10.1145/3528233.3530750} propose to first search for the most plausible motion from the database using the k-Nearest Neighbors (k-NN) algorithm, then refine the motion. DiffGesture~\cite{zhu2023taming} utilizes diffusion models with a cross-modal Transformer network and explores classifier-free guidance to balance diversity and gesture quality.

Nevertheless, co-speech gestures could have a significant inter-person variability due to individual personalities. The aforementioned methods do not explicitly consider speaker identities, necessitating separate models for each speaker and hindering transfer to general scenarios. Furthermore, these methods are limited to modeling either text or audio of speech and fail to combine both modalities. Motivated by these deficiencies, Yoon~\etal~\cite{Yoon2020Speech} propose a generation framework that considers the trimodal context of text, audio, and speaker identity. Bhattacharya~\etal~\cite{bhattacharya2021speech2affectivegestures} further enhance generation quality in terms of affective expressions with an affective encoder and an MFCC encoder. Mix-StAGE~\cite{ahuja2020style} learns unique style embeddings for each speaker and generates motions for multiple speakers simultaneously. HA2G~\cite{Liu_2022_CVPR} employs a hierarchical audio learner to extract audio representations and a hierarchical pose inferer to blend features between audio and body parts. Liu~\etal~\cite{liu2022beat} develop a Cascaded Motion Network (CaMN) that further considers facial expressions, emotions, and semantic meaning based on a large-scale dataset. Rhythmic Gesticulator~\cite{10.1145/3550454.3555435} draws inspiration from linguistic theory and explicitly models both the rhythmic and semantic relations between speech and gestures. TalkSHOW~\cite{yi2022generating} employs an autoencoder for face motions, and a compositional VQ-VAE for body and hand motions based on speech audio and speaker identity. QPGesture~\cite{yang2023QPGesture} introduces a quantization-based and phase-guided motion matching framework using VQ-VAE and Levenshtein distance. LDA~\cite{alexanderson2022listen} demonstrate style control using classifier-free guidance for diffusion models in both music-to-dance, speech-to-gesture, and path-driven locomotion. GestureDiffuCLIP~\cite{Ao2023GestureDiffuCLIP} adapts a latent diffusion model for speech gesture generation and enables controlling with style prompt (text, motion, or video).

\section{Scene-Conditioned Motion Generation}
\label{sec:scene2motion}

Human motion is goal-oriented and influenced by the surrounding scene layout, with individuals moving their bodies to interact with the environment while being constrained by its physical properties. The scene-to-motion generation task aims to generate reasonable human motions consistent with the scene context and has been a long-standing problem in computer graphics and computer vision.
This survey primarily focuses on data-driven scene-conditioned motion generation methods as discussed in Section \ref{sec:scope}, and does not cover methods based on physical simulation~\cite{park2002line, shum2008interaction, agrawal2016task, starke2019neural, holden2017phase}.
Prior to human motion generation, some works have also proposed to synthesize static human poses given a scene condition \cite{gupta20113d, hassan2021populating, savva2016pigraphs, zhang2020place, zhang2020generating}, which will not be further discussed as they also fall outside of the scope of this survey.
In the following, we discuss existing approaches from two perspectives: \textit{scene representation} and \textit{generation pipeline}. Please refer to Table \ref{tab:methods} bottom block.

\subsection{Scene representation}
Current methods exploit various options of scene representations, including 2D images \cite{cao2020long, wang2021scene}, point clouds \cite{mao2022contact, wang2022humanise, huang2023diffusion, araujo2023circle}, mesh \cite{wang2021synthesizing, wang2022towards}, 3D objects \cite{corona2020context, hassan2021stochastic, zhang2022couch, taheri2022goal, wu2022saga, ghosh2022imos} and a specific goal position \cite{zhang2022wanderings, araujo2023circle, huang2023diffusion, wang2021synthesizing}. 
Cao~\etal~\cite{cao2020long} and Wang~\etal~\cite{wang2021scene} use RGB images as the scene constraints which are fused implicitly by extracting features from the image. Many works \cite{mao2022contact, wang2022humanise, huang2023diffusion, araujo2023circle, wang2021synthesizing, wang2022towards} use point clouds or mesh to represent the scene, \eg a room with furniture, and often extract scene features using PointNet \cite{qi2017pointnet} to serve as condition signal. For 3D objects, the configurations include 3D positions of the object \cite{corona2020context, taheri2022goal}, object type \cite{ghosh2022imos, corona2020context} and voxel representation of the object \cite{hassan2021stochastic, zhang2022couch}. For example, Corona~\etal~\cite{corona2020context} represent the object using its 3D bounding box with its object type (\eg, cup) as a one-hot vector, and introduce a directed semantic graph to jointly parameterize the human motion and the object. They use Recurrent Neural Networks (RNN) to generate the human motion to interact with the object. COUCH \cite{zhang2022couch} aims to generate controllable, contact-driven human-chair interactions and represents the chair using an occupancy voxel grid, which accurately captures the spatial relation between the person and the chair. Another typical example using 3D objects as scene conditions involves works that generate whole-body grasping motion \cite{taheri2022goal, wu2022saga, ghosh2022imos}, where 3D object positions \cite{taheri2022goal, ghosh2022imos} or point cloud \cite{wu2022saga} are provided.
Some works give a goal position \cite{wang2021synthesizing, zhang2022wanderings, huang2023diffusion, araujo2023circle} to guide motion generation. For example, GAMMA \cite{zhang2022wanderings} uses Reinforcement Learning to learn a policy network to synthesize plausible motions given the goal position on the ground. SceneDiffuser \cite{huang2023diffusion} proposes a generic framework for diverse 3D scene understanding tasks and uses diffusion models \cite{DDPM} to generate plausible human motions given the point cloud scene and the goal.

Note that most of the methods take more than one type of scene representation as input, and many of them take the past motion or the start pose \cite{corona2020context, cao2020long, wang2021synthesizing, wang2021scene, rempe2021humor, zhang2022couch, mao2022contact, araujo2023circle} as input together.
There also emerge some methods that generate motion with extra language instructions \cite{wang2022humanise} or action labels \cite{hassan2021stochastic, wang2022towards}. For example, HUMANISE~\cite{wang2022humanise} incorporates language descriptions (\eg, walk to the table) to generate human motions in the scene. IMoS \cite{ghosh2022imos} integrates intended action instructions (\eg, drink) to generate a controllable whole-body grasping motion given the object positions and type.

\subsection{Generation pipeline}
Most existing methods propose a multi-stage pipeline. One common pipeline is to first predict the goal position \cite{cao2020long, wang2021scene} or goal interaction anchor \cite{hassan2021stochastic, wang2022towards, zhang2022couch}, then plan a path or trajectory and finally infill the motion along the trajectory \cite{cao2020long, wang2021scene, hassan2021stochastic, wang2022towards, zhang2022couch, mao2022contact, huang2023diffusion}. For example, Cao~\etal~\cite{cao2020long} propose a three-stage motion generation method given the scene as 2D images, which first predicts a 2D goal, then plans a 2D and 3D path, and finally generates the 3D motion along the path via the VAE model. 
Similar to Cao~\etal~\cite{cao2020long}, Wang~\etal~\cite{wang2021scene} use an RGB image as the scene context, and synthesize human future motion by first generating the trajectory and then guiding the motion generation. They further add a discriminator branch to emphasize the consideration of the scene context. 
SAMP \cite{hassan2021stochastic} also adopts a multi-stage pipeline 
which first estimates a goal position and the interaction direction of the object then plans a 3D path given the start body pose, and finally generates reasonable human motions with an autoregressive VAE. 
Compared to SAMP \cite{hassan2021stochastic}, which only models the coarse human-object interaction in the final frame,
Mao~\etal~\cite{mao2022contact} propose to use per-joint contact maps to provide more detailed contact information for every human body joint at each future frame to promote the generation quality. Wang \etal \cite{wang2022towards} first predict diverse human-scene interaction anchors, then incorporate the standard A$^*$ algorithm with scene-aware random exploration for diverse path planning. Finally, a VAE-based framework is used to synthesize anchor poses and complete the motion.
GOAL \cite{taheri2022goal} and SAGA \cite{wu2022saga} aim to generate whole-body grasping motion and propose two-step approaches whereby the ending grasping pose is generated first, followed by the generation of whole-body motion. Different from GOAL \cite{taheri2022goal}, SAGA \cite{wu2022saga} further captures both the diversity of grasping ending poses and the in-between motions by using VAE models.

Some other works utilize given goal positions instead of predicting them. For example, Wang~\etal~\cite{wang2021synthesizing} propose a hierarchical pipeline that uses the VAE model \cite{vae} to generate a static human body on each given sub-goal and generate the in-between human motions for each pair of sub-goal in the scene using bi-directional LSTM \cite{hochreiter1997long}. By stitching these motion clips, long-term human motion is synthesized.
A recent paradigm proposed by CIRCLE \cite{araujo2023circle} is to first initialize the motion using linear interpolation with a given start pose and a goal position, and then propose a scene-aware motion refinement module to generate the final motion. The scene feature is extracted from the 3D point cloud and fused into the refinement module.

\begin{table*}[t]
\center
\caption{Datasets for human motion generation. 
``Kpts.'' and ``Rot.'' denotes keypoints and 3D rotations, respectively.
}
\small
\setlength{\tabcolsep}{5pt}
\resizebox{\linewidth}{!}{
\begin{tabular}{l | c | c | c | c | c | c | c | c | c }
\toprule 
\textbf{Name} & \textbf{Venue} & \textbf{Collection} & \textbf{Representation} & \textbf{Subjects} & \textbf{Sequences} & \textbf{Frames} & \textbf{Length} & \textbf{Condition} & \textbf{Remarks} \\
\toprule 

Human3.6M~\cite{h36m_pami} & TPAMI 2014 & Marker-based & Kpts. (3D) & 11 & - & 3.6M & 5.0h & - & 15 actions \\

CMU Mocap~\cite{cmuWEB} & Online 2015 & Marker-based & Rot. & 109 & 2605 & - & 9h & - & 6 categories, 23 subcategories\\

AMASS~\cite{AMASS:2019} & ICCV 2019 & Marker-based & Rot. & 344 & 11265 & - & 40.0h & - & Unifies 15 marker-based MoCap datasets \\

HuMMan~\cite{cai2022humman} & ECCV 2022 & Markerless & Rot. & 1000 & 400K & 60M & - & - & 500 actions \\

\midrule 

KIT Motion Language~\cite{Plappert2016} & Big data 2016 & Marker-based &  Kpts. (3D) & 111 & 3911 & - & 10.3h & Text & 6.3k Text descriptions \\

UESTC~\cite{ji2019large} & MM 2018 &  Markerless & Kpts. (3D) & 118 & 25.6K & - & 83h & Text & 40 Action classes \\

NTU-RGB+D~\cite{liu2020ntu} & TPAMI 2019 & Markerless & Kpts. (3D) & 106 & 114.4K & - & 74h & Text & 120 Action classes \\

HumanAct12~\cite{guo2020action2motion} & MM 2020 & Markerless & Kpts. (3D) & 12 & 1191 & 90K & 6h & Text & 12 Action classes \\

BABEL~\cite{punnakkal2021babel} & CVPR 2021 & Marker-based  & Rot. & 344 & - & - & 43.5h & Text & 260 Action classes \\

HumanML3D~\cite{Guo_2022_CVPR} & CVPR 2022 &  Marker-based \& Markerless & Kpts. (3D) & 344 & 14.6K  & - & 28.5h & Text & 44.9K Text descriptions \\

\midrule

Tang~\etal~\cite{tang2018dance} & MM 2018 & Marker-based & Kpts. (3D) & - & 61 & 907K & 1.6h & Music & 4 genres \\

Lee~\etal~\cite{NEURIPS2019_7ca57a9f} & NeurIPS 2019 & Pseudo-labeling & Kpts. (2D) & - & 361K & - & 71h & Music & 3 genres \\
 
Huang~\etal~\cite{huang2021} & ICLR 2021 & Pseudo-labeling   & Kpts. (2D) & - & 790 & - & 12h & Music & 3 genres \\
AIST++~\cite{li2021learn} & ICCV 2021 & Markerless & Rot. & 30 & 1,408 & 10.1M & 5.2h & Music & 10 genres \\

PMSD~\cite{Valle-Pérez2021Transflower} & TOG 2021 & Marker-based & Kpts. (3D) & 8 & - & - & 3.8h & Music & 4 genres \\

ShaderMotion~\cite{Valle-Pérez2021Transflower} & TOG 2021 & Marker-based & Kpts. (3D) & 8 & - & - & 10.2h & Music & 2 genres \\

Chen~\etal~\cite{choreomaster2021} & TOG 2021 & Manual annotation & Rot. & - & - & 160K & 9.9h & Music & 9 genres \\

PhantomDance~\cite{DBLP:conf/aaai/LiZZS22} & AAAI 2022 & Manual annotation & Rot. & - & 260 & 795K & 3.7h & Music & 13 genres \\

MMD-ARC~\cite{PC-Dance} & MM 2022 & Manual annotation & Rot. & - & 213 & - & 11.3h & Music & - \\

MDC~\cite{10.1145/3503161.3548090} & MM 2022 & Manual annotation & Rot. & - & 798 & - & 3.5h & Music & 2 genres \\

Aristidou~\etal~\cite{9745335} & TVCG 2022 & Marker-based & Rot. & 32 & - & - & 2.4h & Music & 3 genres \\

AIOZ-GDANCE~\cite{aiozGdance} & CVPR 2023 & Pseudo-labeling & Rot. & >4000 & - & - & 16.7h & Music & 7 dance styles, 16 music genres \\

\hdashline

Trinity~\cite{Trinity2018} & IVA 2018 & Pseudo-labeling & Kpts. (2D) & 1 & 23 & - & 4.1h & Speech & Casual talks \\

TED-Gesture~\cite{yoon2019robots} & ICRA 2019 & Pseudo-labeling & Kpts. (3D) & - & 1,295 & - & 52.7h & Text & TED talks \\

Speech2Gesture~\cite{ginosar2019gestures} & CVPR 2019 & Pseudo-labeling & Kpts. (2D) & 10 & - & - & 144h & Speech & TV shows, Lectures \\

TED-Gesture++~\cite{Yoon2020Speech} & TOG 2020 & Pseudo-labeling & Kpts. (3D) & - & 1,766 & - & 97.0h & Speech, Text & Extension of ~\cite{yoon2019robots} \\

PATS~\cite{ahuja2020style} & ECCV 2020 & Pseudo-labeling & Kpts. (2D) & 25 & - & - & 251h & Speech, Text & Extension of ~\cite{ginosar2019gestures}\\

Speech2Gesture-3D~\cite{habibie2021learning} & IVA 2021 & Pseudo-labeling & Kpts. (3D) & 6 & - & - & 33h & Speech & Videos from ~\cite{ginosar2019gestures}\\

BEAT~\cite{liu2022beat} & ECCV 2022 & Marker-based & Rot. & 30 & 2508 & 30M & 76h & Speech, Text, Emotion & 8 emotions, 4 languages \\

Chinese Gesture~\cite{10.1145/3550454.3555435} & TOG 2022 & Marker-based & Rot. & 5 & - & - & 4h & Speech, Text & Chinese \\

ZEGGS~\cite{ghorbani2022zeroeggs} & CGF 2023 & Marker-based & Rot. & 1 & 67 & - & 2.3h & Speech, Style & 19 Styles  \\

SHOW~\cite{yi2022generating} & CVPR 2023 & Pseudo-labeling & Rot. & - & - & - & 27h & Speech & Videos from ~\cite{ginosar2019gestures}  \\

\midrule

WBHM~\cite{KIT_Dataset} & ICAR 2015 & Marker-based & Rot. & 43 & 3704 & 691K & 7.68h & Object & 41 objects \\
PiGraph~\cite{savva2016pigraphs} & TOG 2016 & Markerless & Kpts. (3D) & 5 & 63 & 0.1M & 2h & Scene, Object & 30 scenes, 19 objects \\
PROX~\cite{hassan2019resolving} & ICCV 2019 & Markerless & Rot. & 20 & 60 & 0.1M & 1h & Scene, Object & 12 indoor scenes \\
i3DB~\cite{iMapper} & SIGGRAPH 2019 & Pseudo-labeling & Kpts. (3D) & 1 & - & - & - & Scene, Object & 15 scenes \\
GTA-IM~\cite{cao2020long} & ECCV 2020 & Marker-based & Kpts. (3D) & 50 & 119 & 1M & - & Scene & Synthetic, 10 indoor scenes \\
GRAB~\cite{taheri2020grab} & ECCV 2020 & Marker-based & Rot. & 10  &  1334  & 1.6M  & - & Object & 51 objects \\
HPS~\cite{HPS} & CVPR 2021 & Marker-based & Rot. & 7 & - & 300K & - & Scene & 8 large scenes, some > 1000 $m^2$ \\
SAMP~\cite{hassan2021stochastic} & ICCV 2021 & Marker-based & Rot. & 1 & - & 185K & 0.83h & Scene, Object & 7 objects \\
COUCH~\cite{zhang2022couch} & ECCV 2022 & Marker-based & Rot. & 6 & >500 & - & 3h & Scene, Chairs & 3 chairs, hand interaction on chairs \\
HUMANISE~\cite{wang2022humanise} & NeurIPS 2022 & Marker-based & Rot. & - & 19.6K & 1.2M & - & Scene, Object, Text & 643 scenes \\
CIRCLE~\cite{araujo2023circle} & CVPR 2023 & Marker-based & Rot. & 5 & >7K & 4.3M & 10h & Scene & 9 scenes \\

\bottomrule 

\end{tabular}
}

\label{tab:dataset}
\end{table*}

\section{Datasets}
 \label{sec:dataset}

In this section, we discuss the datasets for human motion generation. Commonly used datasets can be categorized on the basis of their accompanying conditional signals. We introduce the datasets with paired human motion and conditional signals of text, audio, and scene, respectively. In Table~\ref{tab:dataset}, we summarize the key properties of these datasets and also include the large-scale human motion datasets that do not have extra conditional signals for reference.

\subsection{Text-Motion Datasets}

\noindent\textbf{KIT-Motion Language}~\cite{Plappert2016} is a paired dataset consisting of motion and language data. Motion data are collected via optical marker-based systems, while language data consists of annotations describing each motion datum. 

\noindent\textbf{UESTC}~\cite{ji2019large} includes motion data captured in three modalities - RGB videos, depth and skeleton sequences - using a Microsoft Kinect V2 sensor. The dataset comprises 15 action categories for both standing and sitting position, and 25 categories for standing only, totaling 40 distinct categories.

\noindent\textbf{NTU-RGB+D 120} \cite{liu2020ntu} is an extension of the NTU-RGB+D \cite{shahroudy2016ntu} dataset, with 60 additional classes and 57600 additional RGB+D video samples. The dataset contains 120 different action categories, representing a mix of daily and health-related activities.

\noindent\textbf{HumanAct12}\cite{guo2020action2motion}, derived from the PHSPD \cite{zou20203d}, presents a specialized collection of 3D motion clips, segmented into a spectrum of actions typical of human behavior. The dataset includes daily motions such as walk, run, sit, and warm-up, and is categorized into 12 motion classes and 34 sub-classes.

\noindent\textbf{BABEL}. \emph{Bodies, Action and Behavior with English Labels}\cite{punnakkal2021babel}, provides text labels for motion sequences from the comprehensive motion dataset AMASS\cite{AMASS:2019}. The dataset provides labels on two unique levels: sequence-level for entire sequences and frame-level for individual frames. It covers over 28k sequences and 63k frames across 250 motion categories.

\noindent\textbf{HumanML3D}~\cite{Guo_2022_CVPR} is a dataset derived from the combination of the HumanAct12~\cite{guo2020action2motion} and AMASS~\cite{AMASS:2019} datasets, and it includes three distinct text descriptions corresponding to each motion sequence. The dataset covers a wide range of activities in daily life, sports, acrobatics, and arts.

\subsection{Audio-Motion Datasets}

The audio-motion datasets can be categorized into \emph{controlled} and \emph{in-the-wild} based on the data collection techniques discussed in Section ~\ref{section:motion_data_collection}. The \emph{controlled} audio-motion pairs are obtained by motion capture systems (marker-based, markerless) or manual annotation.
On the contrary, \emph{ in the wild} audio-motion pairs are typically obtained by searching and downloading online videos with specific keywords and utilizing an off-the-shelf pose estimator to extract human motion. Although \emph{in-the-wild} data offer a higher motion diversity and are more scalable, the extracted motions tend to be less accurate.

\subsubsection{Controlled Datasets}

\textbf{Tang~\etal}~\cite{tang2018dance} pioneers to capture 3D dance and corresponding music of 4 types (waltz, tango, cha-cha, and rumba). 

\noindent\textbf{AIST++}~\cite{li2021learn} is constructed from the AIST Dance Video DB~\cite{aist-dance-db}. They utilize multi-view videos to estimate the camera parameters, 3D keypoints, and SMPL parameters. 

\noindent\textbf{PATS.} \emph{Pose-Audio-Transcript-Style}~\cite{Valle-Pérez2021Transflower} dataset consists of synchronized audio and recordings of various dancers and dance styles. 

\noindent\textbf{ShaderMotion}~\cite{Valle-Pérez2021Transflower} extract dance from a social VR platform where the avatars' motion are retargeted from the participants with a 6-point tracking system. 

\noindent\textbf{Aristidou~\etal}~\cite{9745335} invites a group of professional dancers for motion capture and features long sequences of music-dance pairs. 

\noindent\textbf{Trinity}~\cite{Trinity2018} is a multi-modal dataset of conversational speech, containing 4 hours of audio, motion, and video data from one actor. Precise 3D motion is obtained with marker-based motion capture (MoCap) systems. 

\noindent\textbf{BEAT.} \emph{Body-Expression-Audio-Text} dataset~\cite{liu2022beat} is a large-scale semantic and emotional dataset for conversational gestures synthesis, which features rich frame-level emotion and semantic relevance annotations. It also includes facial expressions and multi-lingual speeches. 

\noindent\textbf{Chinese Gesture}\cite{10.1145/3550454.3555435} is a Chinese speech gesture dataset that allows for the exploration of cross-language gesture generation. 

In addition to MoCap-based solutions, several works also propose to extract the audio-motion pairs from character animation resources produced by animators. For example, \textbf{Chen~\etal}~\cite{choreomaster2021} and \textbf{MMD-ARC}~\cite{PC-Dance} utilize MikuMikuDance (MMD) resources from the anime community. \textbf{PhantomDance}~\cite{DBLP:conf/aaai/LiZZS22} recruits a team of experienced animators instructed by professional dancers to create the dance motions. \textbf{MDC.} \emph{Multi-Dancer Choreography}~\cite{10.1145/3503161.3548090} dataset focuses on group dance and they invite the dancers to arrange motion phrases and annotate the temporal dancer activation sequences.

\subsubsection{In-the-wild Datasets}

\textbf{Lee~\etal}~\cite{NEURIPS2019_7ca57a9f} collects dance videos from the Internet with keywords (ballet, Zumba, and hip-hop) and extract 2D body keypoints with OpenPose~\cite{cao2018openpose}. 

\noindent\textbf{Huang~\etal}~\cite{huang2021} addresses the lack of a long-term dance generation dataset. It features one-minute music-dance pairs from the Internet. 

\noindent\textbf{AIOZ-GDANCE}~\cite{aiozGdance} collects in-the-wild group dancing videos along with music and fits SMPL sequences to the tracked 2D keypoints using a temporal extension of SMPLify-X~\cite{pavlakos2019expressive}. They manually fix
the incorrect cases for 2D keypoints and 3D motion, and use human annotations for multi-person relative depth. 

\noindent\textbf{TED-Gesture}~\cite{yoon2019robots} is a co-speech gesture of TED talks that contains videos and English transcripts (along with timestamps for phrases). The authors use OpenPose~\cite{cao2018openpose} to extract 2D poses, then design a neural network to convert 2D poses into 3D poses. 

\noindent\textbf{Speech2Gesture}~\cite{ginosar2019gestures} is a speaker-specific gesture dataset. It is based on the unlabeled in-the-wild videos of television shows and university lectures. The pseudo ground truth is obtained with an off-the-shelf 2D pose estimation algorithm~\cite{cao2018openpose}. The dataset contains $10$ speakers with diverse motion styles, including television show hosts, university lecturers, and televangelists, and therefore enables studying person-specific motion generation. 

\noindent\textbf{TED-Gesture++}~\cite{Yoon2020Speech} extends TED-Gesture~\cite{yoon2019robots} with more videos, featuring synchronized video, speech audio, and transcribed English speech text. The 3D body keypoints are obtained with a temporal 3D pose estimation method~\cite{pavllo20193d}. 

\noindent\textbf{PATS.} \emph{Pose-Audio-Transcript-Style}~\cite{ahuja2020style} extends~\cite{ginosar2019gestures} to more speakers including $15$ talk show hosts, $5$ lecturers, $3$ YouTubers, and $2$ televangelists. Similarly, they extract the skeletal keypoints with OpenPose~\cite{cao2018openpose}. In addition, PATS provides the transcripts corresponding to motion and audio signals. 

\noindent\textbf{Speech2Gesture-3D}~\cite{habibie2021learning} annotates the Speech2Gesture dataset~\cite{ginosar2019gestures} with state-of-the-art 3D face~\cite{GZCVVPT16, saragih2011deformable}, body~\cite{XNect_SIGGRAPH2020} and hand~\cite{zhou2020monocular} pose estimation algorithms. Some videos and subjects from ~\cite{ginosar2019gestures} are excluded due to low resolution and poor 3D reconstruction results. 

\noindent\textbf{SHOW.} \emph{Synchronous Holistic Optimization in the
Wild}~\cite{yi2022generating} fits SMPL-X~\cite{pavlakos2019expressive} parametric model with hand gestures and facial expressions on the Speech2Gesture dataset~\cite{ginosar2019gestures}. It improves SMPLify-X~\cite{pavlakos2019expressive} with advanced regression-based approaches~\cite{feng2021collaborative, feng2021learning, zhang2023pymaf}.

\subsection{Scene-Motion Datasets}

\noindent\textbf{WBHM}. \emph{Whole-Body Human Motion}~\cite{KIT_Dataset} contains 3D whole-body motion data of multiple individuals and objects collected by the Vicon motion capture system. The motion data considers not only the motions of the human subject but the positions and motions of objects with which the subject is interacting as well. $41$ different objects with 3D models are included, such as stairs, cups, food, \etc.

\noindent\textbf{PiGraph}. \emph{Prototypical interaction Graphs} ~\cite{savva2016pigraphs} scans real-world 3D scenes \cite{niessner2013real} and uses Kinect.v2 to capture people's skeletons when they interact with the environments. All objects in the 3D scenes are manually annotated with semantic labels. Multiple interactions are annotated as well.

\noindent\textbf{PROX}. \emph{Proximal Relationships with Object eXclusion}~\cite{hassan2019resolving} contains RGB-D videos of people interacting with real indoor environments, captured by the Kinect-One sensor. The poses of the objects are also captured using attached markers and each object has the CAD model. 

\noindent\textbf{i3DB}~\cite{iMapper} contains several human-scene interactions with annotated object locations and 3D human motion for each captured RGB video. Each object has a class label, such as a chair or table. The 3D human motion is obtained from the estimated 2D motion \cite{Tome_2017_CVPR} with manual corrections.

\noindent\textbf{GTA-IM}. \emph{GTA Indoor Motion}~\cite{cao2020long} is a large-scale synthetic dataset that captures human-scene interactions based on the Grand Theft Auto (GTA) gaming engine. The dataset is equipped with RGB-D videos, 3D human poses, scene instance labels, \etc. Note that the motions in the GTA game engine are from a marker-based motion capture system.

\noindent\textbf{GRAB}. \emph{GRasping Actions with Bodies}~\cite{taheri2020grab} is a large-scale dataset capturing real-world whole-body grasps of 3D objects with Vicon motion capture system. Full-body human motion, object motion, in-hand manipulation, and contact areas are included in the annotation.

\noindent\textbf{HPS}. \emph{Human POSEitioning System}~\cite{HPS} captures 3D humans interacting with large 3D scenes (300-1000 $m^2$, up to 2500 $m^2$), with images captured from a head-mounted camera coupled with the 3D pose and location of the subject in a pre-scanned 3D scene.

\noindent\textbf{SAMP}. \emph{Scene-Aware Motion Prediction}~\cite{hassan2021stochastic} is a rich and diverse human-scene interaction dataset, collected by the high-resolution optical marker MoCap system with $54$ Vicon cameras. Several types of objects such as sofas and armchairs are used during motion capture.

\noindent\textbf{COUCH}~\cite{zhang2022couch} captures humans interacting with chairs in different styles of sitting and free movement. The dataset is collected with IMUs and Kinects and features multiple subjects, real chair geometry, accurately annotated hand contacts, and RGB-D images.

\noindent\textbf{HUMANISE}~\cite{wang2022humanise} is a large-scale and semantic-rich synthetic dataset by aligning the captured human motion sequences in AMASS dataset \cite{mahmood2019amass} with the scanned indoor scenes in ScanNet dataset \cite{dai2017scannet}. Each motion segment has rich semantics about the action type and the corresponding interacting objects, specified by the language description.

\noindent\textbf{CIRCLE}~\cite{araujo2023circle} collects 10 hours of both right and left-hand reaching motion across 9 scenes, captured both in the real world (with the Vicon system) and the VR app. Diverse motions are included such as crawling, bending, \etc.

The above datasets contain not only the scene but also the human motion. Meanwhile, there are also some datasets that only contain scenes and are often used as test sets, such as  \textbf{Matterport3D}~\cite{chang2017matterport3d}, \textbf{Replica}~\cite{straub2019replica}, and \textbf{ScanNet}~\cite{dai2017scannet}.

\section{Evaluation Metrics}
\begin{table*}[t]
    \renewcommand\arraystretch{1.4}
    \centering
    \footnotesize
    \caption{Evaluation metrics for human motion generation. \textcolor{ctext}{Green}, \textcolor{caudio}{blue}, and \textcolor{cscene}{orange} denote \textcolor{ctext}{text-}, \textcolor{caudio}{audio-}, and \textcolor{cscene}{scene-}conditioned motion generation works, respectively.}
    \label{tab:metrics}
    \resizebox{\linewidth}{!}{
    \begin{tabular}{|c|c|p{12 cm}<{\raggedright}|}
        \hline
        \multirow{8}*{\raisebox{0cm}[0pt]{\makecell[c]{Fidelity}}}
        & \multirow{3}*{\raisebox{0cm}[0pt]{\makecell{Comparison with \\ Ground-truth}}}
        & \textbf{Distance}: \ctext{~\cite{8885540, Ghosh_2021_ICCV, petrovich22temos, TEACH:3DV:2022, https://doi.org/10.48550/arxiv.2209.00349, ude2022}}, \caudio{~\cite{tang2018dance, Aud2Repr2Pose, Yoon2020Speech, habibie2021learning, bhattacharya2021speech2affectivegestures, qian2021speech, 10.1145/3550454.3555435, ginosar2019gestures, li2021audio2gestures, DBLP:conf/aaai/LiZZS22, yi2022generating}}, \cscene{~\cite{corona2020context, cao2020long, wang2021synthesizing, rempe2021humor, hassan2021stochastic, taheri2022goal, wu2022saga, mao2022contact, wang2022humanise, huang2023diffusion, ghosh2022imos, araujo2023circle}}\\
        &
        & \textbf{Accuracy}: \caudio{~\cite{ginosar2019gestures, li2021audio2gestures}} \\
        \cline{2-3}
        & \multirow{4}*{\raisebox{0cm}[0pt]{\makecell[c]{Naturalness}}}

        & \textbf{Motion space}: \ctext{~\cite{Ghosh_2021_ICCV, petrovich22temos, TEACH:3DV:2022, https://doi.org/10.48550/arxiv.2209.00349, ude2022}}, \caudio{~\cite{yang2023QPGesture}}, \cscene{~\cite{rempe2021humor, hassan2021stochastic, wang2022towards, wu2022saga}} \\
        &
        & \textbf{Feature space}: \ctext{~\cite{guo2020action2motion, petrovich21actor, cervantes2022implicit, lu2022action, Guo_2022_CVPR, chuan2022tm2t, https://doi.org/10.48550/arxiv.2209.00349, tevet2022human, chen2022mld, ude2022,Lee2023MultiAct, degardin2022generative, zhang2023generating, dabral2022mofusion,yu2020structure}}, \caudio{~\cite{NEURIPS2019_7ca57a9f, huang2021, choreomaster2021, PC-Dance, Yoon2020Speech, bhattacharya2021speech2affectivegestures, qian2021speech, Liu_2022_CVPR, disco2022, liu2022beat, 10.1145/3550454.3555435, zhu2023taming, li2021learn, DBLP:conf/aaai/LiZZS22, siyao2022bailando, Valle-Pérez2021Transflower, ChoreoGraph2022, sun2022you, tseng2022edge, Kim_2022_CVPR, ahuja2020style, 9745335, Ao2023GestureDiffuCLIP, yi2022generating, yang2023QPGesture}}, \cscene{~\cite{wang2021scene, ghosh2022imos}} \\

        \cline{2-3}
        & \multirow{2}*{\raisebox{0cm}[0pt]{\makecell[c]{Physical Plausibility}}}
        & \textbf{Foot sliding}: \cscene{~\cite{taheri2022goal, wu2022saga, araujo2023circle}} \\
        &
        & \textbf{Foot-ground contact}: \caudio{\cite{tseng2022edge}}, \cscene{~\cite{rempe2021humor, zhang2022wanderings, wu2022saga}} \\
        
        \cline{1-3}
        \multirow{6}*{\raisebox{0cm}[0pt]{Diversity}}
        & \multirow{2}*{\raisebox{0cm}[0pt]{\makecell[c]{Intra-motion}}}
        & \textbf{Variation}: \caudio{\cite{li2021audio2gestures, disco2022, 10.1145/3528233.3530750}} \\
        &
        & \textbf{Freezing rate}: \caudio{~\cite{sun2022you}}\\

        \cline{2-3}
        & \multirow{3.5}*{\raisebox{0cm}[0pt]{\makecell[c]{Inter-motion}}}
        & \textbf{Coverage}:  \ctext{~\cite{guo2020action2motion, petrovich21actor, Guo_2022_CVPR, lu2022action, cervantes2022implicit, chuan2022tm2t, tevet2022human, chen2022mld, zhang2023generating, ude2022,Lee2023MultiAct}}, \caudio{~\cite{NEURIPS2019_7ca57a9f, huang2021, choreomaster2021, PC-Dance, Liu_2022_CVPR, zhu2023taming, li2021learn, siyao2022bailando, sun2022you, tseng2022edge, Kim_2022_CVPR, DBLP:conf/aaai/LiZZS22}}, \cscene{\cite{zhang2022couch}} \\
        &
        & \textbf{Multi-modality}: \ctext{~\cite{petrovich21actor, guo2020action2motion, Guo_2022_CVPR, lu2022action, cervantes2022implicit, chuan2022tm2t, tevet2022human, chen2022mld, zhang2023generating, ude2022, Lee2023MultiAct}}, \caudio{~\cite{NEURIPS2019_7ca57a9f, huang2021, li2021audio2gestures}} \cscene{\cite{hassan2021stochastic, wang2022towards, wu2022saga, ghosh2022imos, wang2022humanise, huang2023diffusion}}\\
        
        \hline

        \multirow{6}*{\raisebox{-0.2cm}[0pt]{Condition Consistency}}
        & \multirow{2}*{\raisebox{0cm}[0pt]{\makecell[c]{Text-Motion}}}
        & \textbf{Accuracy}:  \ctext{~\cite{cervantes2022implicit, guo2020action2motion, petrovich21actor, tevet2022motionclip, lu2022action,dabral2022mofusion,lin2023comes, ude2022,zhang2023generating, Guo_2022_CVPR, chuan2022tm2t, chen2022mld, tevet2022human, Lee2023MultiAct}} \\
        &
        & \textbf{Distance}: \ctext{~\cite{zhang2023generating, Guo_2022_CVPR, chuan2022tm2t, chen2022mld, tevet2022human,lin2023comes, https://doi.org/10.48550/arxiv.2209.00349}}\\

        \cline{2-3}
        & \multirow{3}*{\raisebox{0cm}[0pt]{\makecell[c]{Audio-Motion}}}
        & \textbf{Beat}:  \caudio{\cite{NEURIPS2019_7ca57a9f, huang2021, 10.1145/3550454.3555435, li2021learn, Kim_2022_CVPR, sun2022you, aiozGdance, disco2022, liu2022beat, siyao2022bailando, DBLP:conf/aaai/LiZZS22, ChoreoGraph2022, tseng2022edge, Liu_2022_CVPR, zhu2023taming, yang2023QPGesture}} \\
        &
        & \textbf{Semantics}: \caudio{\cite{liu2022beat, Ao2023GestureDiffuCLIP}} \\

        \cline{2-3}
        & \multirow{2}*{\raisebox{0cm}[0pt]{\makecell[c]{Scene-Motion}}}
        & \textbf{Non-collison score}: \cscene{\cite{wang2021synthesizing, wang2021scene, hassan2021stochastic, wang2022towards, taheri2022goal, wu2022saga, huang2023diffusion, araujo2023circle}}  \\
        &
        & \textbf{Human-scene contact}: \cscene{\cite{wang2021synthesizing, wang2022towards, taheri2022goal, wu2022saga, zhang2022couch, huang2023diffusion}} \\
        \hline
        \multirow{3}*{\raisebox{-0.2cm}[0pt]{User Study}}
        & \multirow{3.5}*{\raisebox{0cm}[0pt]{\makecell[c]{Subjective Evaluation}}} & \textbf{Preference}: 
        \ctext{\cite{guo2020action2motion, Guo_2022_CVPR, chuan2022tm2t, tevet2022human, petrovich22temos, 8885540}},
        \caudio{\cite{Ao2023GestureDiffuCLIP, alexanderson2022listen, yi2022generating, qian2021speech, liu2022beat, disco2022, Yoon2020Speech, ahuja2020style, NEURIPS2019_7ca57a9f, huang2021, li2021learn, DBLP:conf/aaai/LiZZS22, siyao2022bailando, Kim_2022_CVPR, sun2022you, 9745335, tseng2022edge}}, 
        \cscene{\cite{ghosh2022imos}}
        \\
        & & \textbf{Rating}:  
        \ctext{\cite{hong2022avatarclip}},
        \caudio{\cite{yang2023QPGesture, zhu2023taming, 10.1145/3550454.3555435, 10.1145/3528233.3530750, Liu_2022_CVPR, li2021audio2gestures, bhattacharya2021speech2affectivegestures, habibie2021learning, takeuchi2017creating, ginosar2019gestures, tang2018dance, Valle-Pérez2021Transflower, choreomaster2021, PC-Dance, ChoreoGraph2022, 10.1145/3503161.3548090, 9745335}}, 
        \cscene{\cite{wang2021synthesizing, wang2021scene, wang2022towards, taheri2022goal, wu2022saga, zhang2022wanderings, wang2022humanise}} 
        \\
        \hline  

    \end{tabular}
    }
    \vspace{-2mm}
\end{table*}
 \label{sec:metrics}

Proper evaluation metrics are vital to compare different methods and drive the progress of the field. However, evaluation of synthesized human motion is a non-trivial problem due to the one-to-many mapping nature, human evaluation subjectivity, and high-level cues of the conditional signals. In this section, we summarize the commonly-used evaluation metrics from different aspects, and discuss their strengths and limitations. See Table \ref{tab:metrics} for a summary.

\subsection{Fidelity}

The fidelity metrics aim to evaluate the general quality of the generated motions in terms of naturalness, smoothness, plausibility, \etc.

\subsubsection{Comparison with Ground-truth}

In assessing the quality of generated motion, comparing it to the ground truth serves as the most straightforward approach. 

\noindent\textbf{Distance.} Most works~\cite{8885540, Ghosh_2021_ICCV, petrovich22temos, TEACH:3DV:2022, https://doi.org/10.48550/arxiv.2209.00349, ude2022, tang2018dance, Aud2Repr2Pose, Yoon2020Speech, habibie2021learning, bhattacharya2021speech2affectivegestures, ginosar2019gestures, li2021audio2gestures, qian2021speech, 10.1145/3550454.3555435, corona2020context, cao2020long, wang2021synthesizing, rempe2021humor, hassan2021stochastic, taheri2022goal, wu2022saga, mao2022contact, wang2022humanise, huang2023diffusion, ghosh2022imos, araujo2023circle} employ the distance metrics to measure the difference between synthesized motion and ground truth motion. Li~\etal~\cite{DBLP:conf/aaai/LiZZS22} utilize Normalized Power Spectrum Similarity (NPSS)~\cite{gopalakrishnan2019neural} for evaluating long-term motion synthesis capabilities. \revision{NPSS operates in the frequency domain and is less sensitive to frame misalignment compared to MSE. Meanwhile, Normalized Directional Motion Similarity (NDMS) \cite{tanke2021intention} is proposed to measure the similarity of the motion direction and the ratio of motion magnitudes in the motion prediction field.}

\noindent\textbf{Accuracy.} As direct distance computation alone might not provide a thorough evaluation, some works~\cite{ginosar2019gestures, li2021audio2gestures} further compute the Percentage of Correct 3D Keypoints (PCK)~\cite{yang2012articulated} which has been a popular evaluation metric of pose estimation. To compute PCK, the proportion of accurately generated joints is determined, with a joint deemed accurate if its distance to the target remains within a predefined threshold.

However, the ground truth represents only one feasible outcome for the given conditional input, with countless alternative solutions being potentially adequate. Consequently, relying solely on ground truth comparisons for motion generation evaluation may lack comprehensive coverage.

\subsubsection{Naturalness}

Motion quality evaluates the naturalness of the generated motion, which is usually measured by comparing the generated motion manifold with the real motion manifold. Existing metrics can be categorized into \emph{motion space} and \emph{feature space}, based on the space used for evaluation.

\noindent\textbf{Motion space.} Some approaches measure the distribution distance based on geometric statistics in the motion space. For example, some works~\cite{Ghosh_2021_ICCV, petrovich22temos, TEACH:3DV:2022, https://doi.org/10.48550/arxiv.2209.00349, ude2022} report Average Variance Error (AVE) which compute the difference between the variance of the real motion and the synthesized motion. 
QPGesture~\cite{yang2023QPGesture} measures the Hellinger distance~\cite{kucherenko2020gesticulator} between the speed-distribution histograms. Some works~\cite{rempe2021humor, yang2023QPGesture} also compares the higher-order derivatives of joint positions (acceleration, jerk).
SAMP~\cite{hassan2021stochastic} and Wang~\etal~\cite{wang2022towards} calculate the Fréchet distance (FD) of the two distributions based on the pose rotations. 
\revision{In motion prediction literature, Power Spectrum Entropy (PSEnt) and KL divergence (PSKL) \cite{hernandez2019human} are used for computing the distribution distance. SAGA \cite{wu2022saga} utilizes PSKL-J \cite{hernandez2019human,zhang2021learning} to measure the acceleration distribution of generated and real motion to evaluate motion smoothness.}

\noindent\textbf{Feature space.} The second category is to compute the distribution distance in the feature space using a standalone neural network as motion feature extractor.
To this end, some works compute Fr\'echet Inception Distance (FID) using an auxiliary action classifier~\cite{NEURIPS2019_7ca57a9f, huang2021, ghosh2022imos, petrovich21actor,lu2022action, Guo_2022_CVPR, chuan2022tm2t, https://doi.org/10.48550/arxiv.2209.00349, tevet2022human, chen2022mld, Lee2023MultiAct, zhang2023generating, dabral2022mofusion} or an autoencoder~\cite{choreomaster2021, PC-Dance, Yoon2020Speech, bhattacharya2021speech2affectivegestures, qian2021speech, Liu_2022_CVPR, disco2022, liu2022beat, 10.1145/3550454.3555435, zhu2023taming, wang2021scene, Ao2023GestureDiffuCLIP, yang2023QPGesture, alexanderson2022listen}. The metric can be extended by disentangling the motion feature into two aspects of geometric (pose) and kinetic (movement)~\cite{li2021learn, DBLP:conf/aaai/LiZZS22, siyao2022bailando, Valle-Pérez2021Transflower, ChoreoGraph2022, sun2022you, tseng2022edge, alexanderson2022listen}. These works take advantage of the well-designed motion feature extractors ~\cite{mullerefficient2005, onuma2008fmdistance, gopinath2020fairmotion, Guo_2022_CVPR, guo2020action2motion} to calculate the feature distance. Kim~\etal~\cite{Kim_2022_CVPR} further train a dance genre classifier to extract the style features and calculate corresponding FID. 
Except for FID, several other metrics are employed to compute the distribution distance between generation and real, including Inception Score (IS)~\cite{salimans2016improved, ahuja2020style}, chi-square distance~\cite{9745335}, Maximum Mean Discrepancy (MMD)~\cite{yu2020structure, degardin2022generative}, Mean Maximum Similarity(MMS)~\cite{cervantes2022implicit}, Canonical correlation analysis (CCA)~\cite{sadoughi2019speech, yang2023QPGesture}, and realistic score~\cite{yi2022generating}.

\revision{Although these metrics are intuitive, there exist several critical challenges. Their evaluation of naturalness highly depends on dataset distribution and the effectiveness of the pretrained motion feature extractor, which may not be comprehensive to reflect overall motion quality. For instance, EDGE~\cite{tseng2022edge} illustrates that the prevailing FID score is inconsistent with human evaluations, questioning the effectiveness of the common practice.}

\subsubsection{Physical Plausibility}
\label{subsubsec:physical}
Physical plausibility refers to the degree to which a generated motion is in accordance with the physical rules, particularly relevant to foot-ground interactions: (1) foot sliding, and (2) foot-ground contact. 

\noindent\textbf{Foot sliding.} Some work \cite{taheri2022goal, wu2022saga, araujo2023circle} 
measure the foot skating artifacts of the generated motion. For example, SAGA \cite{wu2022saga} defines skating as when the heel is within a threshold of the ground and the heel speed of both feet exceeds a threshold. CIRCLE \cite{araujo2023circle} reports the percentage of frames in a sequence with foot sliding.

\noindent\textbf{Foot-ground contact.} Previous work has proposed several different metrics. For example, EDGE~\cite{tseng2022edge} reports the physical foot contact score (PFC).  SAGA \cite{wu2022saga} reports a non-collision score which is defined as the number of body mesh vertices above the ground divided by the total number of vertices. HuMoR \cite{rempe2021humor} reports the binary classification accuracy of person-ground contacts and the frequency of foot-floor penetrations of the generated motion. GAMMA \cite{zhang2022wanderings} computes the contact score by setting a threshold height from the ground plane and a speed threshold for skating.
\revision{However, at present, there is a lack of a standardized metric for quantifying physical plausibility. Various methods may employ disparate parameter choices and even design distinct evaluation approaches. Consequently, there is a potential requirement for the development of a more robust and universally applicable metric that effectively measures the degree of physical plausibility.}

\subsection{Diversity}

Another important goal is to generate various human motions and avoid repetitive contents. To this end, researchers measure the generation results from different levels: diversity within single motion sequence (intra-motion diversity) and diversity among different motion sequences (inter-motion) diversity.

\subsubsection{Intra-motion Diversity}
Long-sequence motion generation tends to incur the ``freezing'' problem~\cite{huang2021, sun2022you}. To evaluate ``non-freezability'' and discriminate the static motions, some works measure the intra-motion diversity metrics.

\noindent\textbf{Variation.} For example, some studies~\cite{li2021audio2gestures, disco2022} split the generated motions into equal-lengthed non-overlapping motion clips and calculate their average pairwise distance. Habibie~\etal~\cite{10.1145/3528233.3530750} measure the temporal position and velocity variations.

\noindent\textbf{Freezing rate.} Sun~\etal~\cite{sun2022you} propose to calculate the temporal differences of the pose and translation parameters and report a freezing rate. 

\subsubsection{Inter-motion Diversity}

To evaluate the inter-motion diversity of the generated motion manifold, the existing metrics can be categorized into \emph{coverage} and \emph{multi-modality}.

\noindent\textbf{Coverage.} The coverage of generated motion manifold is usually evaluated by first sampling $N$ different conditional signals on the validation set, then computing the diversity of the generated motions. For example, ~\cite{NEURIPS2019_7ca57a9f, huang2021, choreomaster2021, PC-Dance, Liu_2022_CVPR, zhu2023taming, yang2023QPGesture, aiozGdance, guo2020action2motion, petrovich21actor, Guo_2022_CVPR, lu2022action, cervantes2022implicit, chuan2022tm2t, tevet2022human, chen2022mld, zhang2023generating, ude2022, Lee2023MultiAct} reports the average feature distance of the model results. Similar to FID, the feature distance can be divided into geometric, kinetic~\cite{li2021learn, siyao2022bailando, sun2022you, tseng2022edge}, and style~\cite{Kim_2022_CVPR}. Some works~\cite{DBLP:conf/aaai/LiZZS22, yang2023QPGesture, zhang2022couch} also calculate diversity in the motion space.

\noindent\textbf{Multi-modality.} Given the same conditional signal, the probabilistic generative methods could generate a distribution over the plausible corresponding motions. The multi-modality metrics aim to evaluate the variations of the distribution. The common practice is to first sample $N$ different conditional signals on the validation set, then generate $M$ motions for each condition, and calculate the average pairwise distance for each condition. Existing methods report the average feature distance~\cite{NEURIPS2019_7ca57a9f, huang2021, wu2022saga, ghosh2022imos, hassan2021stochastic, yi2022generating, petrovich21actor, guo2020action2motion, Guo_2022_CVPR, lu2022action, cervantes2022implicit, chuan2022tm2t, tevet2022human, chen2022mld, zhang2023generating, ude2022,Lee2023MultiAct} or average pose distance~\cite{li2021audio2gestures, wang2022towards, hassan2021stochastic, wang2022humanise, huang2023diffusion}. ODMO~\cite{lu2022action} also uses normalized APD (n-APD)~\cite{yuan2020dlow}, which is determined by the ratio of APD values between generated motions and ground truth. \revision{Yuan \etal \cite{yuan2020dlow} also utilize average displacement error (ADE), final displacement error (FDE), multi-modal ADE (MMADE), and multi-modal FDE (MMFDE) based on the multi-modal nature of the problem in motion prediction.} Some work further evaluates the generation diversity at levels of interaction anchors or planned paths~\cite{wang2022towards}.

\subsection{Condition Consistency}

The above metrics all focus on the properties of the generated motion itself, while it is also crucial to evaluate the consistency between the generated motion and the corresponding conditional signals. As these evaluation metrics highly correlate with the condition types, we will discuss them according to different tasks.

\subsubsection{Text-Motion Consistency}

\noindent\textbf{Accuracy.} In assessing the consistency between the generated motions and the corresponding texts in action-to-motion tasks, various existing approaches leverage recognition accuracy~\cite{cervantes2022implicit, guo2020action2motion, petrovich21actor, tevet2022motionclip, lu2022action,Lee2023MultiAct} to evaluate the generation results. This metric is based on a pretrained action recognition model and determines whether the generated motions can be correctly identified as their corresponding action classes. The use of recognition accuracy provides a high-level view of how well the generated samples fit within the expected action class, given the textual description. In addition, some methods ~\cite{dabral2022mofusion,lin2023comes, ude2022,zhang2023generating, Guo_2022_CVPR, chuan2022tm2t, chen2022mld, tevet2022human} use R-Precision to assess the correspondence between the generated motions and their associated descriptions. This metric calculates and ranks the Euclidean distances among the features and averages the accuracy of the top-k results, offering a granular measure of text-motion consistency.

\noindent\textbf{Distance.} On the other hand, some methods delve deeper into the feature-level distance to measure the text-motion consistency. For instance, Multimodal Distance~\cite{zhang2023generating, Guo_2022_CVPR, chuan2022tm2t, chen2022mld, tevet2022human} quantifies the disparity between the feature from a given description and the motion feature from the generated result, providing a direct measure of the feature-level alignment between the text and the motion. Similarly, Motion CLIP Score (mCLIP)~\cite{lin2023comes, https://doi.org/10.48550/arxiv.2209.00349} utilizes cosine similarity to capture the closeness between text features and motion features in CLIP space, providing a quantifiable measure of how well the modalities align. Flame~\cite{https://doi.org/10.48550/arxiv.2209.00349} further leverage Mutual Information Divergence (MID)~\cite{kim2022mutual} to measure the alignment between different modalities.

\revision{Nonetheless, these metrics are significantly influenced by the performance of the pretrained models, as well as the quality and distribution of data used for their training. Consequently, these metrics may have limitations in their ability to offer an objective evaluation.}

\subsubsection{Audio-Motion Consistency}

\noindent\textbf{Beat.} Existing methods typically assess the degree to which kinematic beats of generated motions align with the input audio beats. To achieve this, beat coverage and hit rate~\cite{NEURIPS2019_7ca57a9f, huang2021, 10.1145/3550454.3555435} represents the ratio of aligned beats to all beats. Li~\etal~\cite{li2021learn} proposes a beat alignment score calculated with beat distances and is followed by ~\cite{Kim_2022_CVPR, sun2022you, aiozGdance, disco2022, liu2022beat}. Some later works~\cite{siyao2022bailando, DBLP:conf/aaai/LiZZS22, ChoreoGraph2022, tseng2022edge, yang2023QPGesture, alexanderson2022listen} further refine the score definition by emphasizing music beat matching. In addition, studies~\cite{Liu_2022_CVPR, zhu2023taming} suggest using mean angle velocity instead of position velocity. 

\noindent\textbf{Semantics.} To further evaluate the semantic consistency, Liu~\etal~\cite{liu2022beat} propose Semantic-Relevant Gesture Recall (SRGR), which weighs PCK based on semantic scores of the ground truth data. They suppose that it is more in line with subjective human perception than the L1 variance. GestureDiffuCLIP~\cite{Ao2023GestureDiffuCLIP} proposes semantic score (SC) to measure the semantic similarity between generated motion and transcripts in their joint embedding space.

\revision{At present, most evaluation metrics primarily focus on the basic connection between audio and motion, often neglecting subtler and cultural connections like style and emotion. For instance, hip-hop music and ballet dance may not be considered harmonious by human standards even with well-aligned beats. The same applies to a speech in a sad tone accompanied by cheerful gestures. Unfortunately, these nuances have not been fully addressed by existing audio-motion consistency metrics.}

\subsubsection{Scene-Motion Consistency}
\label{subsubsec:scene_consistency}
We distinguish physical plausibility (Section \ref{subsubsec:physical}) and scene-motion consistency by dividing the scene into the ground and other objects. 
The scene-motion consistency refers to the agreement of the generated motion with the given scene condition (except for ground). There are mainly two perspectives to evaluate the consistency: (1) non-collision score, and (2) human-scene contact.

\noindent\textbf{Non-collision score} is a metric used to evaluate the safety and physical plausibility of a generated motion colliding with other objects or obstacles in the environment \cite{wang2021synthesizing, wang2021scene, hassan2021stochastic, wang2022towards, taheri2022goal, wu2022saga, huang2023diffusion, araujo2023circle}. For example, Wang \etal \cite{wang2021scene} compute human-scene collision as the intersection points between a human motion represented as a cylinder model and the point cloud of the given scenes. The non-collision ratio is defined as the ratio between the number of human motions without human-scene collision and all sampled motions. Some work \cite{wu2022saga, taheri2022goal} uses body-scene penetration for this metric. For example, SAGA \cite{wu2022saga} measures the interpenetration volumes between the body and object mesh, and GOAL \cite{taheri2022goal} reports the volume of the penetrations ($cm^3$).

\noindent\textbf{Human-scene contact} focuses on the contact areas to evaluate the scene-motion consistency \cite{wang2021synthesizing, wang2022towards, taheri2022goal, zhang2022couch, wu2022saga, huang2023diffusion}, and has different definitions considering different scene conditions. SAGA \cite{wu2022saga} measures the ratio of body meshes being in minimal contact with object meshes to evaluate the grasp stability. COUCH \cite{zhang2022couch} focuses on how well the synthesized motion meets the given contacts by using Average Contact Error (ACE) as the mean squared error between the predicted hand contact and the corresponding given contact, as well as Average Contact Precision (AP$@k$), where a contact is considered as correctly predicted if it is closer than $k$ $cm$.

There are some other metrics that aim to evaluate how well the generated motion reaches the final goal state, such as the execution time \cite{hassan2021stochastic}, the success rate of the character reaching the goal within several attempts \cite{zhang2022wanderings}, the body-to-goal distance \cite{zhang2022wanderings, wang2022humanise, araujo2023circle}. The execution time \cite{hassan2021stochastic} is the time required to transition to the target action label from an idle state. HUMANISE \cite{wang2022humanise} and CIRCLE \cite{araujo2023circle} assess the body-to-goal distance to evaluate how well the generated motion interacts with or reaches the correct object. \revision{In summary, various novel metrics have been proposed to measure scene-motion consistency, reflecting the diverse and intricate nature of scene representations. However, these metrics are often specifically crafted to address unique aspects within their respective research contexts, potentially restricting their broad generality and universal applicability.}

\subsection{User Study}

User study, or subjective evaluation, serves as an essential component in evaluating generated motions, as it can uncover aspects of motion quality that may not be captured by objective metrics alone. Firstly, humans are highly sensitive to minor artifacts in biological motion, such as jittering and foot skating~\cite{troje2002decomposing, shimada2012modulation}. Secondly, current objective metrics are unable to encompass nuanced cultural aspects of generated motions, \eg, aesthetics and emotional impact. Existing methods design user studies that focus on one or several of the aforementioned aspects (quality, diversity, consistency) using \emph{preference} or \emph{rating}.

\noindent\textbf{Preference.} 
Many studies employ user study with pairwise preference comparisons between their generation results and baselines or GT. Specifically, participants observe a pair of human motions and respond to questions such as, "Which motion corresponds better to the textual description?", "Which dance is more realistic, regardless of music?", "Which dance matches the music better in terms of style?", or "Which motion best satisfies scene constraints?", \etc. Subsequently, researchers calculate a win rate for their method against the baselines. The preference-based user studies offer a direct evaluation between the compared methods; however, they may be insufficient for comparing multiple methods. To address this, EDGE~\cite{tseng2022edge} performs pairwise comparisons among all generation methods and uses the Elo Rating~\cite{elo1978rating} to represent their generation quality simultaneously.

\noindent\textbf{Rating.} 
Another prevalent user study approach involves instructing volunteers to provide explicit scores for the generation results. Participants are typically shown multiple motion generations and asked to assign a score (\eg, from 1 to 5) for each motion. Some studies further require a separate score for each aspect (quality, diversity, consistency).

\section{Conclusion and Future Work}
 \label{sec:conclusion}

In this survey, we provide a comprehensive overview of recent advancements in human motion generation. We begin by examining the fundamental aspects of this problem, specifically focusing on human motion and generation methods. Subsequently, we classify research studies based on their conditional signals and discuss each category in detail. Furthermore, we provide a summary of available dataset resources and commonly used evaluation metrics. Despite the rapid progress in this field, significant challenges remain that warrant future exploration. In light of this, we outline several promising future directions from various perspectives, hoping to inspire new breakthroughs in human motion generation research.

\noindent\textbf{{Data.}} 
 Different from images or video, collecting high-quality human motion data is much more difficult and expensive, which leads to a trade-off in data quantity and data quality. Furthermore, the variability in motion representations and conditional signals hinders the broad applicability of existing datasets. To address these issues, future research could investigate the use of heterogeneous data sources, integrating their benefits through weakly-supervised learning approaches~\cite{motionbert2022, li2023isolated} or multi-modal foundation models~\cite{tevet2022motionclip, xue2023ulip}.

\noindent\textbf{{Semantics.}} 
 It is worth noting that human motion is more than just the movement of body parts; it also serves as a crucial non-verbal communication tool, conveying semantic information within cultural and societal contexts. Capturing the semantic relationship between human motion and conditional signals (\eg high-level text descriptions, music/speech styles, and environment affordances) is essential for visually appealing and aesthetically pleasing results that align with human perception. 
A specific challenge in this field is how to equip generative models with prior knowledge of human motion semantics. Some studies~\cite{tevet2022motionclip, Ao2023GestureDiffuCLIP} adopt the pretrained foundation models with language priors.
We believe that future research could delve deeper into exploring semantic connections from various perspectives, encompassing data, methodology, and evaluation.

\noindent\textbf{{Evaluation.}} 
 As discussed in Section ~\ref{sec:metrics}, appropriate evaluation metrics for human motion are crucial yet challenging. Although various objective evaluation metrics have been explored, they all possess inherent limitations and cannot supplant subjective user studies~\cite{tseng2022edge}. Future work could focus on devising more principled objective evaluation metrics that not only align closely with human perception but also maintain interpretability.

\noindent\textbf{{Controllability.}} 
 The ability to control the generation content is important in real-world applications, which has been a popular topic in image generative models~\cite{li2019controllable,Shoshan_2021_ICCV,mokady2023null}. Some recent works explore controllable human motion generation with joint mask~\cite{tseng2022edge} or style prompt~\cite{Ao2023GestureDiffuCLIP}. We believe that future works could further explore controllability to create more user-friendly experiences, \eg, interactive and fine-grained editing~\cite{pan2023drag}.

\noindent\textbf{{Interactivity.}} 
 The interactive nature of human motion is important but has not been fully explored yet. Most current studies focus primarily on generating single-human motion within static environments. Future works could delve into human motion generation in the context of human-human and human-environment interactions. Examples of potential areas of exploration include motion generation for closely interacting social groups (\eg, conversation, group dance, \etc) and motion generation in dynamic, actionable scenes~\cite{habitat19iccv, szot2021habitat, li2022behavior}.

\bibliographystyle{IEEEtran}
\bibliography{sample}

\end{document}